%% file: main.tex
\documentclass[10pt,twocolumn,letterpaper,table]{article}

\usepackage{3dv}
\usepackage{times}
\usepackage{epsfig}
\usepackage{graphicx}
\usepackage{amsmath}
\usepackage{amssymb}
\usepackage{tikz}
\usetikzlibrary{backgrounds,fit}

\include{includes}

\usepackage[pagebackref=true,breaklinks=true,letterpaper=true,colorlinks,bookmarks=false]{hyperref}

\threedvfinalcopy

\def\threedvPaperID{6182} 

\ifthreedvfinal\fi
\begin{document}

\title{Web Stereo Video Supervision for Depth Prediction from Dynamic Scenes}

\author{Chaoyang Wang~~~~~Simon Lucey\\
Carnegie Mellon University\\
{\tt\small \{chaoyanw, slucey\}@cs.cmu.edu}
\and
Federico Perazzi~~~~~Oliver Wang\\
Adobe Inc.\\
{\tt\small \{perazzi, owang\}@adobe.com}
}

\twocolumn[{
\renewcommand\twocolumn[1][]{#1}%
\maketitle
\begin{center}
    \centering
    \includegraphics[width=.95\linewidth]{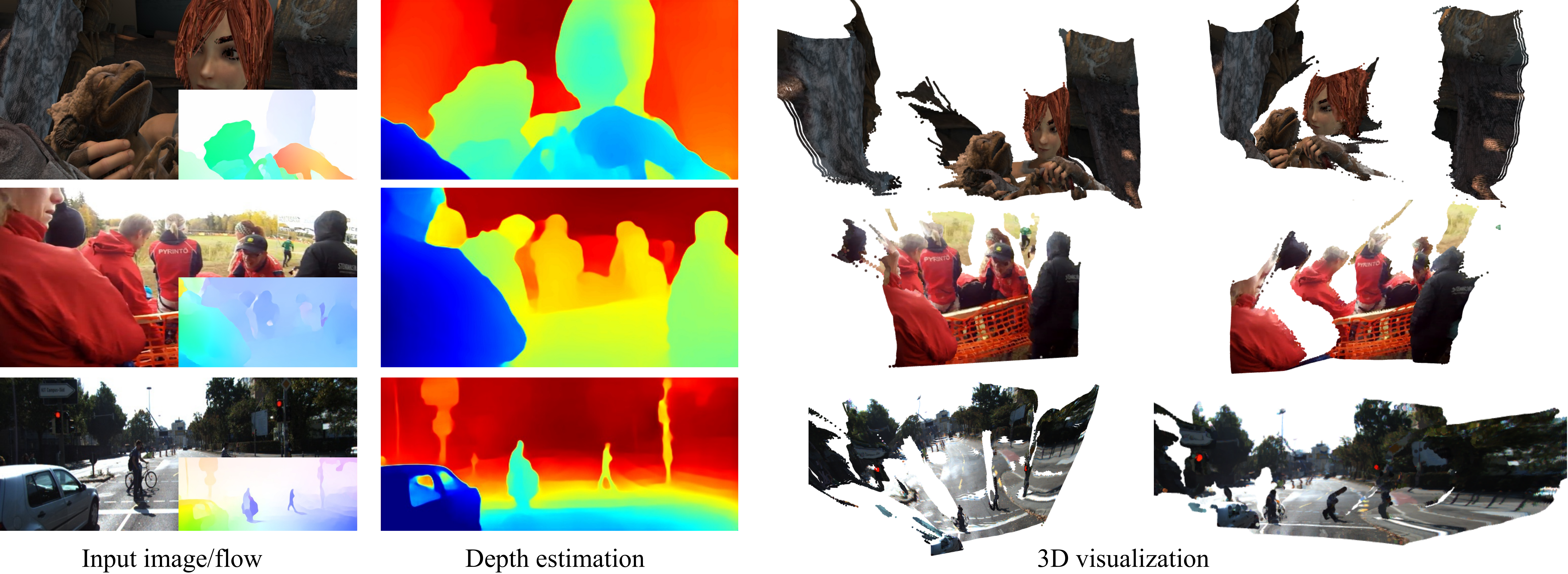}
    \captionof{figure}{Depth prediction for nonrigid scenes from our multi-view depth estimator, which is trained on a new large scale database of web stereo videos.}
\end{center}
}]

\input{sections/abstract.tex}

\input{sections/introduction.tex}

\input{sections/relatedwork.tex}

\input{sections/dataset.tex}

\input{sections/approach.tex}

\input{sections/evaluation.tex}

\input{sections/discussion.tex}

{\small
\bibliographystyle{ieee}
\bibliography{wsvd_ref}
}


\end{document}


\title{Web Stereo Video Supervision for Depth Prediction from Dynamic Scenes \\ Supplementary Material}

\author{First Author\\
Institution1\\
Institution1 address\\
{\tt\small firstauthor@i1.org}
\and
Second Author\\
Institution2\\
First line of institution2 address\\
{\tt\small secondauthor@i2.org}
}

\maketitle

\input{sections/appendix.tex}

{\small
\bibliographystyle{ieee}
\bibliography{nonrigiddepth}
}


%% file: includes.tex
\usepackage{url}
\usepackage{amsthm}
\usepackage{color}
\usepackage{subcaption}
\usepackage{booktabs}
\usepackage{multirow}
\usepackage{pbox}
\usepackage[demo,abs]{overpic}
\usepackage{enumitem}

\usepackage{colortbl}
\usepackage{booktabs}
\usepackage{xfrac}
\usepackage{tablefootnote}
\usepackage{tabularx}
\definecolor{rowblue}{RGB}{220,230,240}

\usepackage{etoolbox}
\makeatletter
\patchcmd\@combinedblfloats{\box\@outputbox}{\unvbox\@outputbox}{}{\errmessage{\noexpand patch failed}}
\makeatother

\newcommand{\dataset}{\textsc{WSVD}}

%% file: sections/abstract.tex
\begin{abstract}

We present a fully data-driven method to compute depth from diverse monocular video sequences that contain large amounts of non-rigid objects, e.g., people.
In order to learn reconstruction cues for non-rigid scenes, we introduce a new dataset consisting of stereo videos scraped in-the-wild.
This dataset has a wide variety of scene types, and features large amounts of nonrigid objects, especially people. 
From this, we compute disparity maps to be used as supervision to train our approach. 
We propose a loss function that allows us to generate a depth prediction even with unknown camera intrinsics and stereo baselines in the dataset. 
We validate the use of large amounts of Internet video by evaluating our method on existing video datasets with depth supervision, including SINTEL, and KITTI, and show that our approach generalizes better to natural scenes.


\end{abstract}

%% file: sections/introduction.tex
\section{Introduction}
Recovering depth maps of non-rigid scenes from monocular video sequences is a challenging problem in 3D vision. 
While rigid scene reconstruction methods can use geometric consistency assumptions across multiple frames, the problem becomes severely under-constrained for nonrigid scenes.
As a result, nonrigid reconstruction methods must rely more heavily on strong scene priors.
In fact, we see that data-driven single image reconstruction methods, which can learn \emph{only} scene priors, sometimes outperform multi-frame geometric methods on nonrigid scenes (Sec.~\ref{sec:eval}). 
In this work, we attempt the model nonrigid scene priors in a \emph{data-driven} manner, by training on real-life sequences, while also learning to taking advantage of geometric cues between neighboring frames of video. 

Recent advances in data-driven methods have shown some advantages over traditional 3D reconstruction pipelines.
However, due to restrictions on architectures and available training data, such approaches have largely been used to predict depth from from single images~\cite{eigen2014depth,li2018megadepth,xian2018monocular,diw,laina2016deeper} or multiple views of rigid scenes~\cite{ummenhofer2017demon,schoenberger2016mvs}, or have been trained on narrow domains, e.g, driving data~\cite{geiger2012we}. 


To overcome these limitations, we introduce a new  large-scale (1.5M frame) dataset collected in-the-wild from internet stereo videos, which contains large amounts of non-rigid objects and diverse scene types. We use derived stereo disparity for training, and at test time, predict depth from a pair of single-view, sequential frames. 
Our approach is designed so that it learns to utilize both semantic (single-image) and geometric (multi-frame) cues to address the nonrigid reconstruction problem.


One challenge of using Internet stereo videos as a training source is that they contain unknown (and possibly temporally varying) camera intrinsics and extrinsics, and consequently, we cannot directly translate disparities to depth values (even up to scale). This prevents the use of existing regression loss including the scale invariant logarithmic depth/gradient loss~\cite{eigen2014depth}.
Instead, we observe that in a stereo camera configuration, the \emph{difference} in disparity between two pixels is proportional to the difference in their inverse depth.
This motivates a new normalized multiscale gradient loss that allows for supervision of depth prediction networks directly from estimated disparities from uncalibrated stereo video data.
Compared to the ordinal loss used in ~\cite{diw,xian2018monocular}, we show that our proposed loss has the advantage of retaining distance information between points, and yields smoother, more accurate depth maps. 

While the computed disparity maps contain errors, we show that training a deep network on our dataset with the proposed loss generalizes well to other datasets, and that by using temporal information in the form of flow maps and sequential frames as input into the network, we are able to improve over single image depth prediction methods. 
Our code and data will be made available to the research community.

In summary, we present the following contributions: 
\begin{enumerate}[topsep=0pt,itemsep=-1ex,partopsep=1ex,parsep=1ex]
    \item A simple architecture that leverages stereo video for training, and produces depth maps from monocular sequential frames with nonrigid objects at test time. To our knowledge this is the first multi-frame data-driven approach that outperforms single-image reconstruction on nonrigid scenes.
    \item A new stereo video dataset that features a wide variety of real world examples with non-rigid objects (e.g., humans) in an unconstrained setting. 
    \item A novel loss function that allows us to train with unknown camera parameters, while outperforming previously used ordinal losses. 
\end{enumerate}

%% file: sections/relatedwork.tex
\section{Related Work}

\noindent\textbf{Traditional geometric approaches} for predicting depth from video sequences rely on SLAM~\cite{mur2015orb} or SfM~\cite{schoenberger2016sfm} pipelines for camera pose and 3D map estimation, followed by dense multiview stereo~\cite{schoenberger2016mvs} to get per-pixel depth values.
These methods are mature systems that produce highly accurate reconstructions in the right circumstances.
However, they rely on hand-designed features, assumptions such as brightness constancy across frames, and often require a good map initialization and sufficient parallax to converge. 
Additionally, while camera pose estimation can handle some amount of nonrigid motion by simply ignoring those regions, dense multiview stereo requires a rigid scene.

Non-rigid SfM methods try to recover 3D by replacing the rigid assumption with additional constraints. 
Bregler et al.~\cite{bregler2000recovering} introduces a fixed rank constraint on the shape matrix for non-rigid structures. 
Numerous innovations have followed, introducing additional priors to make the problem less ambiguous
and to scale to dense reconstruction\cite{kong2016prior,dai2014simple,lee2013procrustean,zhu2014complex,akhter2009nonrigid,kumar2018scalable}. 
These approaches usually assume weak perspective cameras and rely on hand-designed features for input feature tracks. 

Other recent works\cite{kumar2017monocular,dmde} explore dense reconstructions from two perspective frames, using an as-rigid-as-possible (ARAP) assumption to address the scale ambiguity issue inherent in non-rigid reconstruction. 
Although promising results have been shown on mostly rigid scenes, their ARAP assumption is not enough to handle complicated dynamic scenes. 
Our approach learns priors from data, and we show that it can often produce good result for highly dynamic scenes.

\noindent\textbf{Data driven approaches} that leverage deep networks for scene priors and feature learning, have become a potential way to overcome the limitations of traditional methods. 
Recent works~\cite{ummenhofer2017demon, huang2018deepmvs,zhou2018deeptam} demonstrate promising result for rigid scene reconstruction. 
However, since they include explicit rigid transformation inside their network architecture, they cannot handle non-rigid motion. 
We propose a new data-driven method that focuses on non-rigid objects and diverse scenes, and introduce a new internet stereo video dataset to train this approach.

\noindent\textbf{Supervision using depth sensors}
is a common strategy for existing methods that directly regress depth values.
These approaches rely on datasets collected for example, by laser scanner~\cite{saxena2009make3d}, Kinect depth camera~\cite{eigen2014depth}, car-mounted lidar sensor~\cite{eigen2014depth,kuznietsov2017semi}, synthetic rendered data (for pretraining)~\cite{Guo_2018_ECCV}, or dual-camera iPhone~\cite{deeplens2018}.
One challenge of requiring specialized hardware is that it is hard to acquire sufficiently diverse data for training, and as such they tend to be used only in constrained domains, e.g., driving sequences. \\
Recent approaches have proposed using more common stereo camera rigs for training depth prediction networks~\cite{godard2017unsupervised,yang2018lego,zhan2018unsupervised}.
These approaches are trained using stereo video data on KITTI~\cite{geiger2012we}, by treating depth prediction as an image reconstruction problem.
Our approach differs to theirs in that we are learning from stereoscopic videos with unknown camera parameters, and we also utilize temporal information at test time.
Another method, Deep3D~\cite{xie2016deep3d} uses 3D movies for training, however this approach focuses on synthesizing novel stereoscopic views rather than scene reconstruction.

\noindent\textbf{Supervision using internet images}
is a powerful tool that allows for the collection of diverse datasets for learning depth reconstruction priors.
MegaDepth~\cite{li2018megadepth} generates a set of 3D reconstructions using traditional SfM and multi-view stereo reconstruction pipelines from internet images.
These reconstructions serve as ground truth normalized depth maps for supervision.
Another recent work scrapes stereo \emph{images} from the web, and computes disparity from these~\cite{xian2018monocular}. 
This approach uses ordinal depth constraints from these disparity estimates to train a single image depth prediction network.


In contrast, we are interested in data that allows us to take advantage of multiple frames of video.
We show that reconstruction quality is improved when such data is available, indicating that it is possible to learn both scene priors from single images, and geometric information from multiple frames.
In addition, we compare our loss formulation to that presented in the above method, and show that it outperforms relative depth constraints. 

\noindent\textbf{Supervision using single-view video data} has been a popular recent approach for scene reconstruction.
In this case, supervision is derived from temporal frames of a video by predicting depth and camera pose, and computing a loss based on warping the one frame to the next based on these values. 
Then at test time, a single frame is used for depth prediction~\cite{garg2016unsupervised,MahjourianWA17,zou2018dfnet,zhou2018stereo,zhou2017unsupervised,mahjourian2018unsupervised,Wang_2018_CVPR,yang2018lego}. 
As geometric projection is only valid for rigid scenes, these methods often include a rigidity mask, where the loss is treated differently outside these regions~\cite{zhou2017unsupervised,zou2018dfnet}, and introduce regularization to prevent the method from degenerating. 


While these approaches are elegant as they do not require extra supervision other than a single-view video, in practice many of these methods require both known intrinsics, and assume mostly rigid scenes.
In our work we are interested in highly non-rigid scenes, and therefore do not explicitly rely on a geometric reprojection loss.
Instead we use temporal flow and flow-warped images as an input to our network to learn the geometric relationship between flow and depth.


\begin{figure}
    \centering
    \includegraphics[width=\linewidth]{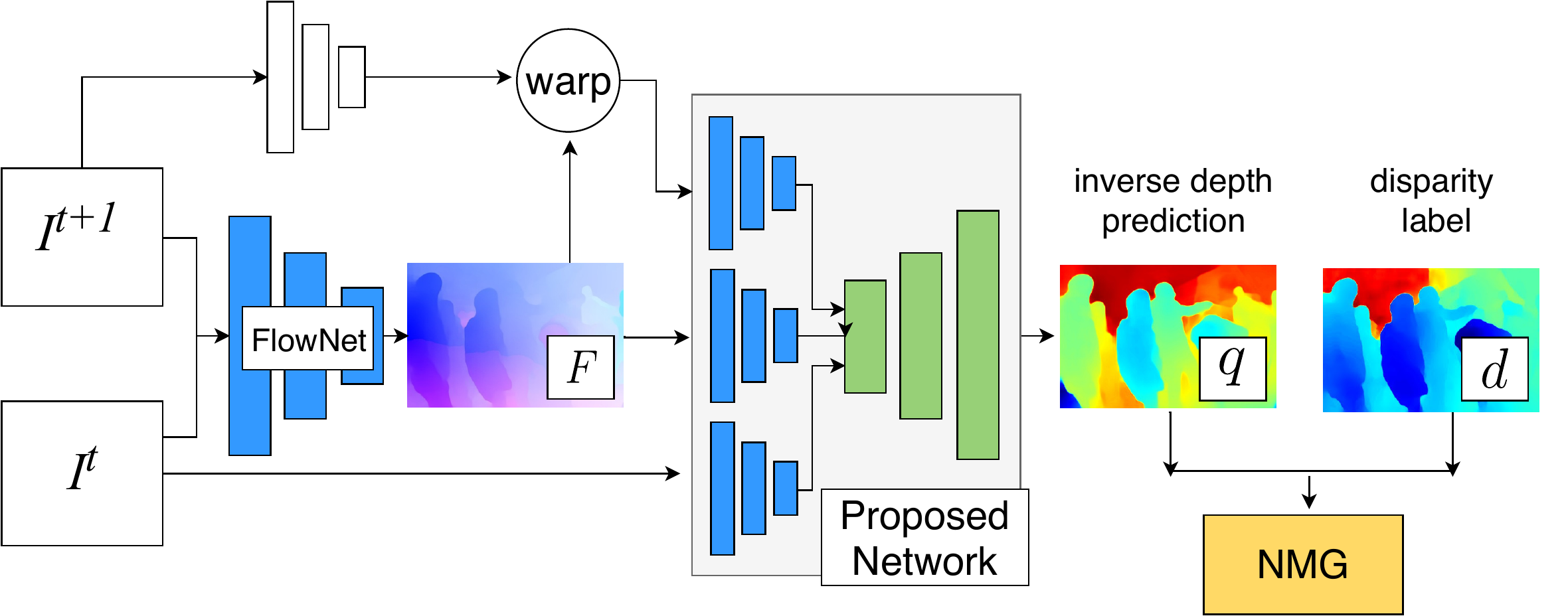}
    \caption{Our network takes input from: the frame $I^t$, the second frame $I^{t+1}$ and the flow between $I^t,I^{t+1}$. We extract feature pyramids for both input images and the flow map. Moreover, we warp the feature pyramid for $I^{t+1}$ with the flow. The warped feature pyramid of $I^{t+1}$ is then fused with the feature pyramids for $I^t$ and the flow map, fed into a decoder with skip connections and produces a depth map. This is supervised with the disparity directly using our Normalized Multi-Scale Gradient (NMG) Loss.}
\end{figure}

%% file: sections/dataset.tex
\begin{figure}[t]
    \centering
    \begin{tabular}{*{4}{c@{\hspace{2px}}}}
    \includegraphics[width=.22\linewidth]{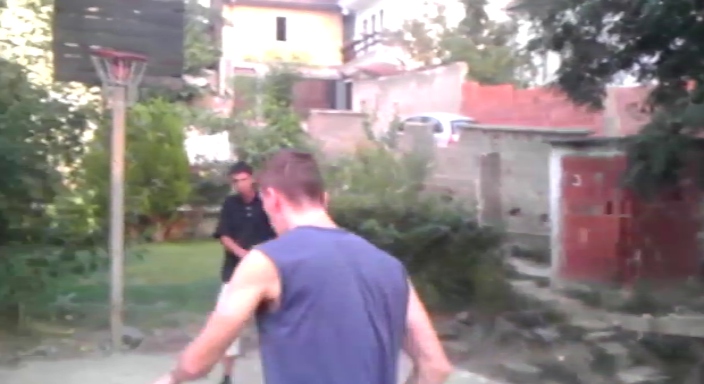} &
    \includegraphics[width=.22\linewidth]{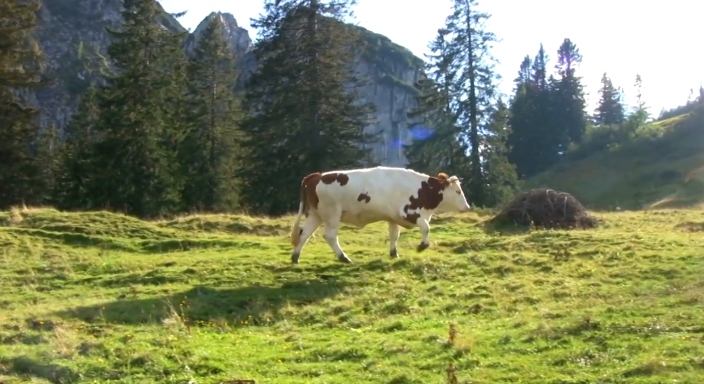} &
    \includegraphics[width=.22\linewidth]{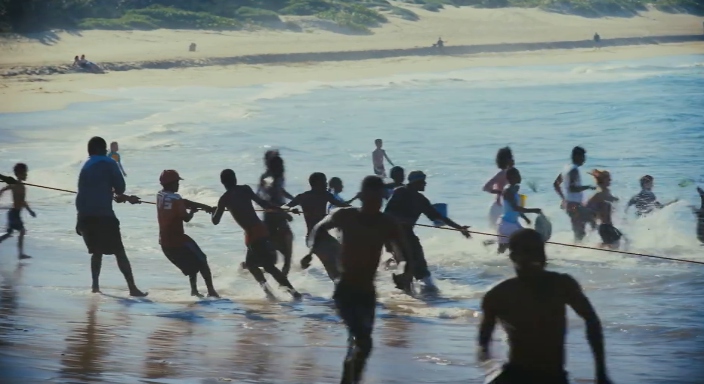} &
    \includegraphics[width=.22\linewidth]{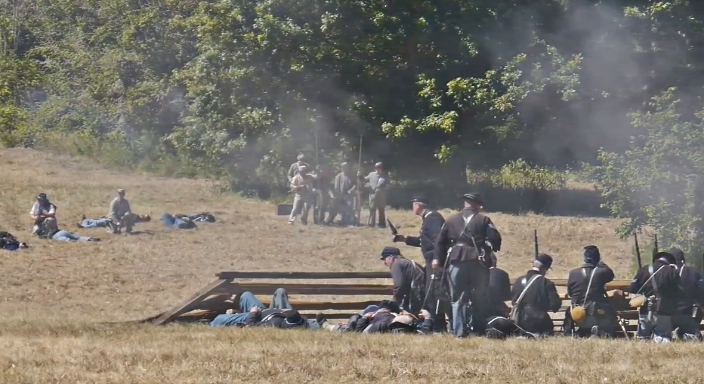} \\
    
    \includegraphics[width=.22\linewidth]{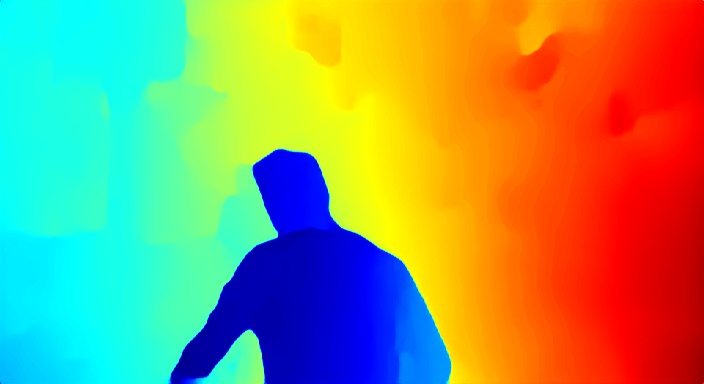} &
    \includegraphics[width=.22\linewidth]{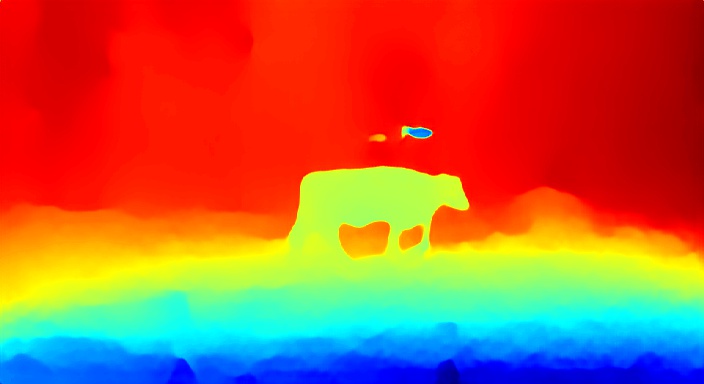} &
    \includegraphics[width=.22\linewidth]{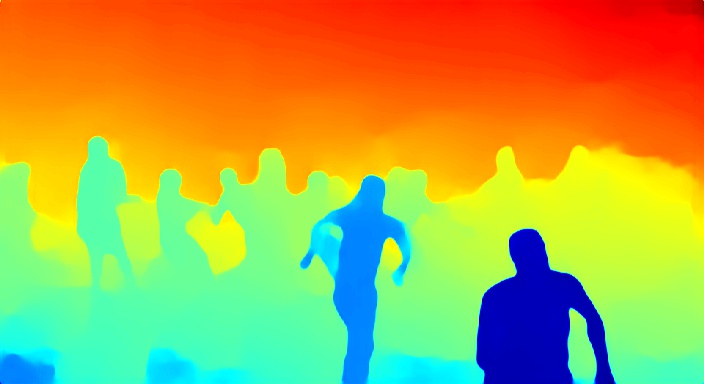} &
    \includegraphics[width=.22\linewidth]{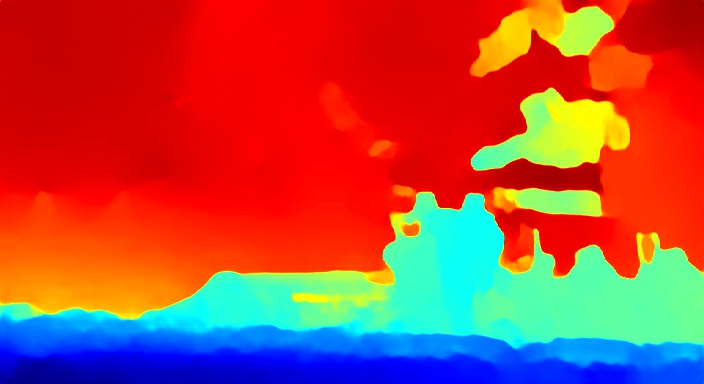} \\

    \includegraphics[width=.22\linewidth]{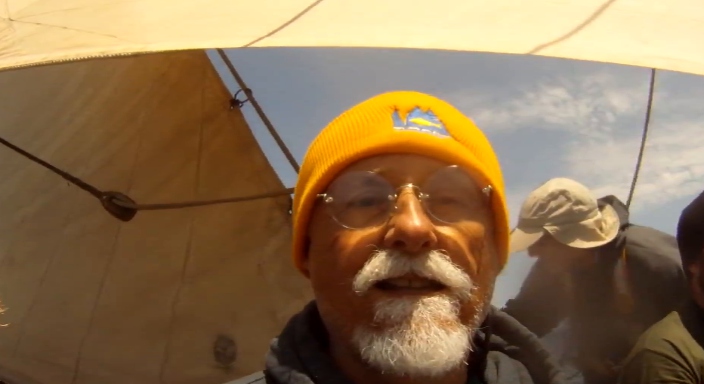} &
    \includegraphics[width=.22\linewidth]{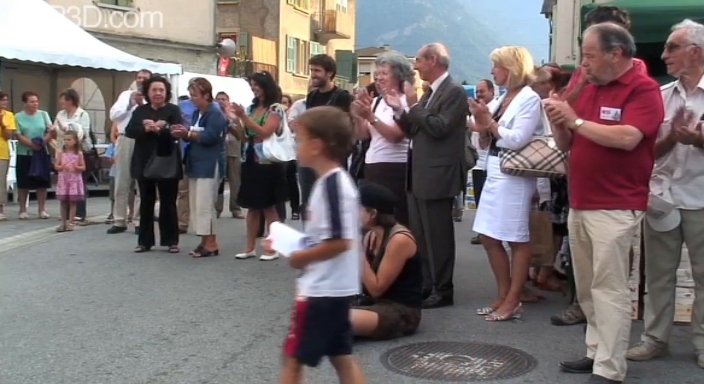} &
    \includegraphics[width=.22\linewidth]{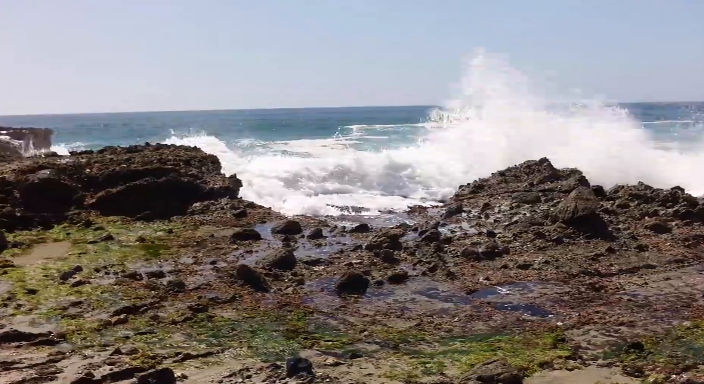} &
    \includegraphics[width=.22\linewidth]{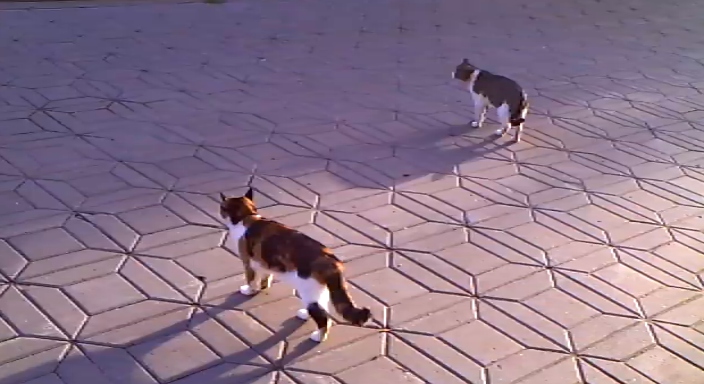} \\
    
    \includegraphics[width=.22\linewidth]{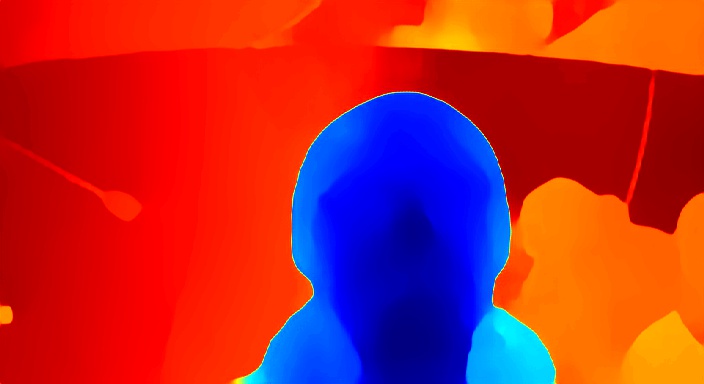} &
    \includegraphics[width=.22\linewidth]{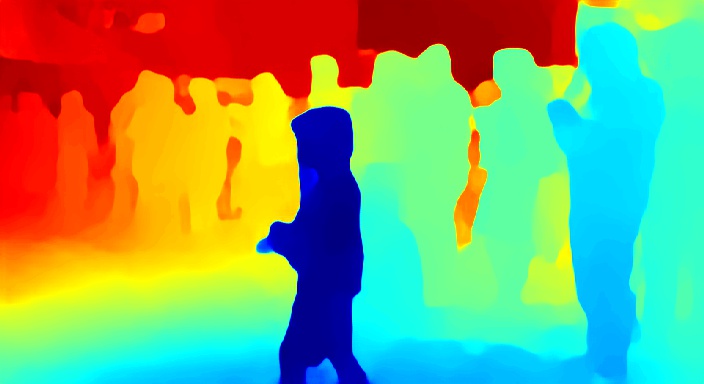} &
    \includegraphics[width=.22\linewidth]{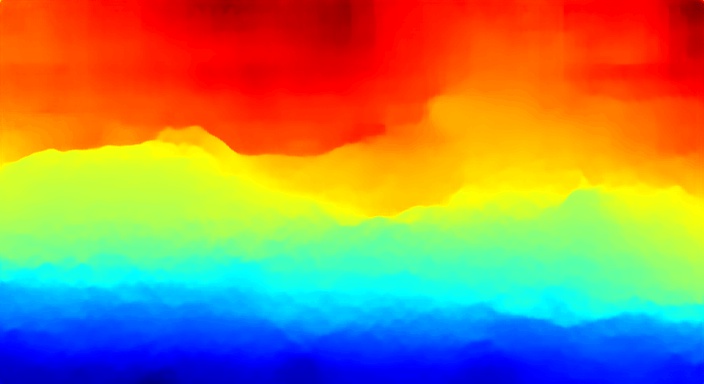} &
    \includegraphics[width=.22\linewidth]{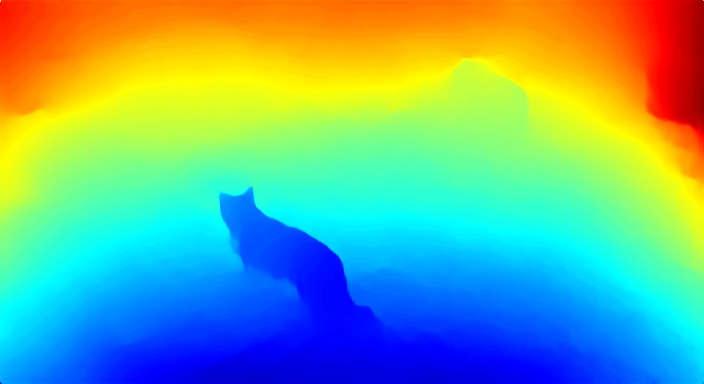} \\
    
    \end{tabular}   
    \caption{Random selection of frames from \dataset{} showing a sampling of the diversity of scenes and subjects contained. Below each image we show the computed disparity map used for supervision.}
    \label{fig:dataset}
\end{figure}

\section{Dataset}
We introduce the Web Stereo Video Dataset (\dataset{}), a large, in-the-wild collection of stereo videos featuring many non-rigid objects (especially people).
Figure~\ref{fig:dataset} shows a selection of representative frames from \dataset{}.

To collect \dataset{}, we scraped YouTube for videos that were tagged as being viewable in stereo 3D.
3D videos on YouTube are stored internally in a left/right format, meaning that each video is squeezed in half width-wise and stored side by side in a single video file.
We additionally searched for videos on Vimeo with a ``side-by-side 3D'' query to match this format (Vimeo does not naively support 3D).
In theory all videos with the proper tag should be 3D, but in practice, a large number of these videos were either regular monocular video, or different kinds of stereo, e.g., anaglyph, top/bottom, etc. 
We therefore conduct two stages of filtering to make this data usable.

\paragraph{Filtering}
Our initial scraping yielded $7k+$ videos from YouTube and Vimeo.
We first identified all videos that were actual left-right stereo videos from this set.
To do this, we computed the MSE of the left and right half of each video, split the videos into two classes based on on this metric and then manually removed outliers. 

After this step, all videos remaining were left/right stereo videos, but still not all of them were usable. For further filtering, we split up each video into shots using histogram differences \cite{pyscendetect}, and performed a per-shot categorization of good and bad videos. To do this, we first calculated the average brightness of the middle frame to filter out shots with black screens. We also performed text detection to remove shots with large text titles.

We then automatically computed the disparity for the middle frames per shot using a flow approach~\cite{ilg2017flownet}. We found that many samples displayed substantial lens distortion, near-zero baselines, vertical disparities, color differences, inverted cameras, and other poor stereo characteristics. To reject those shots, we used the following criteria as an initial step for filtering:
pixels with vertical disparity $>$ 1 pixel is greater than $10\%$;
range of horizontal disparity is $<$ 5 pixels, and the percentage of pixels passing a left-right disparity consistency check is $<$ $70\%$.
This was followed by a second curation step, to remove videos with obviously incorrect disparity maps (e.g., due to radial distortion, or incorrect camera configurations).

Next, we removed static frames by filtering frames with maximum flow magnitude less than 8 pixels.
For each remaining frame, we calculated disparity map as our ground truth using FlowNet2.0~\cite{ilg2017flownet}, which we found to produce the best results. 
The left-right disparity check is also used to mask out outliers, which are not used for supervision. 
Finally, we are left with 10788 clips from 689 videos, consisting of around 1.5 million frames.
\begin{figure}
    \centering
    \includegraphics[width=.8\linewidth]{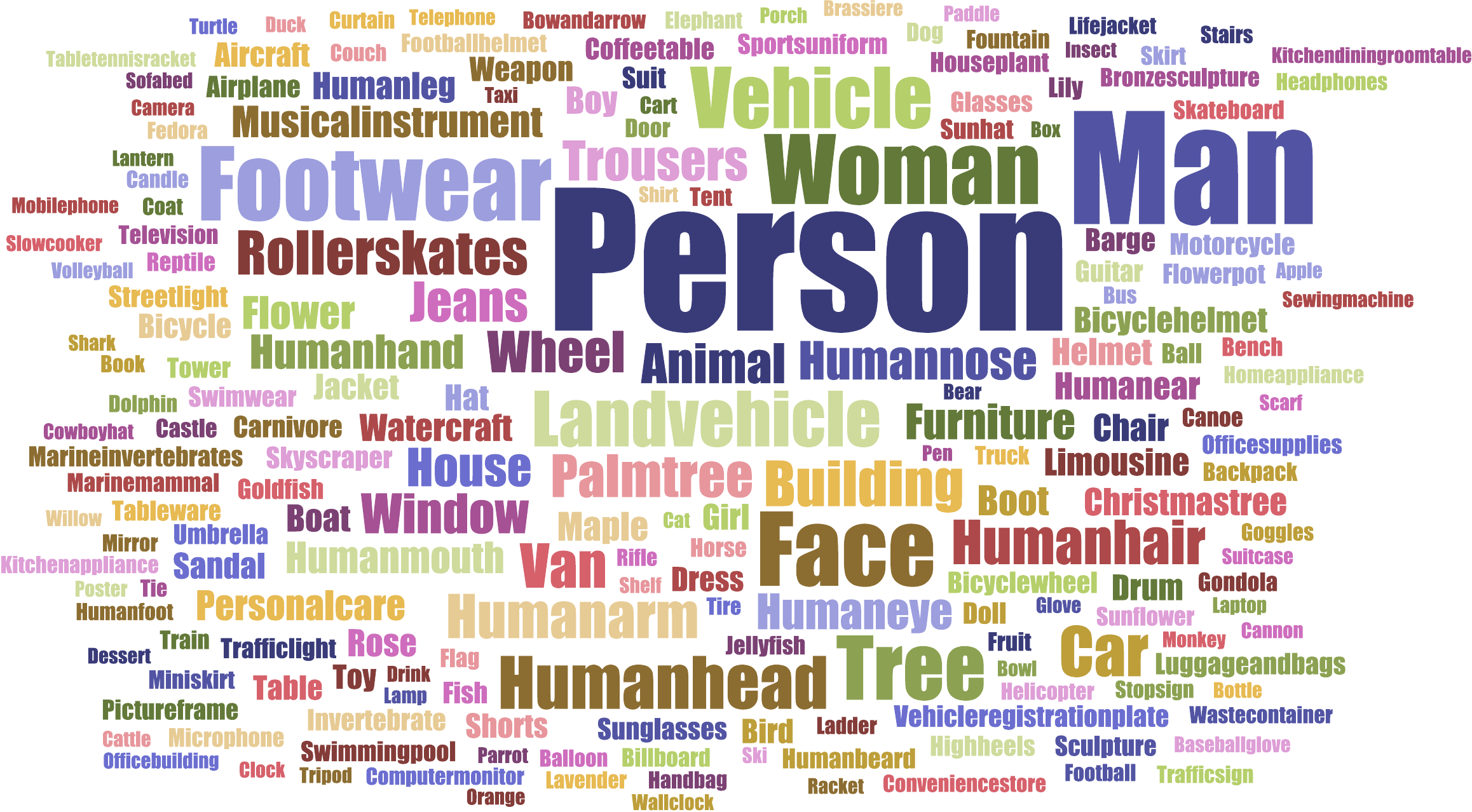}
    \caption{\dataset{} word cloud of the two hundred most frequent classes estimated with an object detector trained on OpenImages~\cite{openimages} bounding boxes, showing a high quantity of non-rigid objects, especially people.}
    \label{fig:wordcloud}
\end{figure}

\paragraph{Analysis}
In order to understand what type of content is present in the \dataset{} dataset, we run a two-stage image classifier similar to Mask R-CNN \cite{He2017MaskR} on the middle frames of each shot. The classifier is pre-trained on the 600 classes annotated in the Open Images dataset~\cite{openimages}. 
We retained bounding boxes with confidence score $\geq0.7$. 
The analysis of the results indicates that roughly $79\%$ of the video sequences contain either humans, animals, or vehicles, which are likely to display non-rigid motion. In Figure~\ref{fig:wordcloud}, we show the word cloud generated from the class frequencies of the detected bounding boxes. In contrast to other datasets, i.e. KITTI and NYU, \dataset{} reflects the diverse content that people watch on YouTube, with many non-rigid objects, especially humans.

%% file: sections/approach.tex
\section{Approach}\label{sec:approach}

Our supervision comes in the form of a disparity map.
Assuming that the stereoscopic videos are captured by two horizontally placed cameras, and the vertical disparity has already been removed, then disparity can be translated to inverse depth $q$ using the following equation:
\begin{equation}
    q = \frac{d-(c^R_x - c^L_x)}{fb}
\label{eq:inv_depth}
\end{equation}
Where $d$ is the disparity, $b$ is the camera baseline, $f$ is the focal length, and $c^R_x$, $c^L_x$ denote the horizontal coordinates of the principal points for the left/right cameras. $c^R_x - c^L_x$ defines the minimum possible disparity $d_\text{min}$ in the camera configuration. In practice, $d_\text{min}$ can have arbitrary values depending on the stereo rig configuration and post-production. For example, to release visual fatigue, most stereo rigs for 3D movies are towed-in to place the object of interest at the 0 disparity plane, which creates negative disparity values.

Unlike most other video datasets with depth supervision, this dataset has unknown and variable focal length $f$, and camera stereo configuration (baseline $b$ and $d_\text{min}$). 
Among those,  $d_\text{min}$ is the key parameter preventing us from converting the estimated disparity into an inverse depth map up to scale, as is commonly done in other self-supervised learning methods~\cite{godard2017unsupervised,li2018megadepth}. This term also prevents us to apply the widely used scale-invariant logarithmic depth (gradient) loss~\cite{eigen2014depth}, due to the fact that subtracting the mean/neighboring pixel's logarithmic depth value is not enough to cancel out $d_\text{min}$.

Although in theory, we could estimate $d_\text{min}$ with the disparity value of pixels at infinity distance, it would not be robust due to the fact that regions in distance does not always present in videos, and that the usual choice for such regions i.e. textureless sky, are likely to have incorrect disparity values.

Due to this issue, prior work~\cite{xian2018monocular} only uses an ordinal relation for supervision, and does not attempt to recover the relative distance from the disparity map. We take a different approach -- the proposed loss function takes supervision from the whole disparity map, and preserves the continuous distance information between points in addition to ordinal relation. In comparison, ordinal loss~\cite{xian2018monocular,diw} only enforce binary ordinal relation between a sparse set of pixel pairs.

\subsection{Normalized multiscale gradient (NMG) loss}
From Eq.~\ref{eq:inv_depth} we derive that the difference in disparity between two pixels is proportional to the difference in their inverse depth:
\begin{equation}
    q_i-q_j = \frac{d_i-d_j}{fb}
\end{equation}
This allows us to design a novel loss invariant to the minimum possible disparity value.
The idea is to enforce the gradient of inverse depth prediction to be close to the gradient of the disparity map up to scale (normalized).
The gradient is evaluated at different spacing amounts (multiscale) to include both local/global information~\cite{ummenhofer2017demon}. The loss can be written as:
\begin{equation}
    \mathcal{L} = \sum_{k}\sum_{i} |s\nabla_x^k  q_i - \nabla_x^k  d_i | + |s\nabla_y^k  q_i - \nabla_y^k  d_i |,
\end{equation}
where $\nabla_x^k$, $\nabla_y^k$ denote the difference evaluated with spacing $k$ (we use $k = \{2,8,32,64\}$); and the scale ratio $s$ is estimated by:
\begin{equation}
    s = \frac{\sum_k\sum_i |\nabla_x^k d_i| + \sum_k\sum_i |\nabla_y^k d_i|}{\sum_k\sum_i |\nabla_x^k q_i| + \sum_k\sum_i |\nabla_y^k q_i| }.
\end{equation}
We choose to scale the inverse depth prediction to match disparity, and not the other way around, as this is more robust when the signal-to-noise ratio of the disparity map is low, which often happens when the range of disparity values is narrow, either due to a small baseline, or distant scenes. 
Scaling such noisy disparity with low contrast would amplify noise in the depth prediction.

Compared to the scale-invariant gradient loss of Eigen et al.~\cite{eigen2014depth}, the proposed loss NMG is different in that it is defined in terms of disparity, not the log of depth. which is inapplicable in our scenario since we do not have depth as supervision. 

Compared to the ordinal loss~\cite{xian2018monocular,diw}, our NMG loss enforces relative distance and smoothness in addition to pairwise ordinal relation.
As shown in Figure~\ref{fig:ordinal}, the NMG loss yields more accurate global structure and preserves edge details better.

\subsection{Depth prediction network}
To utilize temporal information for depth prediction, we propose a network architecture inspired by recent work~\cite{xian2018monocular}, modified to take input from two sequential frames and their optical flow.
The general architecture is a multi-scale feature fusion~\cite{zou2018dfnet,xian2018monocular} net with three feature pyramids streams (features for 1st image, warped features for 2nd frame, and features for the flow map).
These features are then projected and concatenated together and fed into an decoder with skip connections. Please refer to the supplementary material for a detailed description of the architecture.

\vspace{5px} \noindent \textbf{Feature pyramids} 
We use a ResNet-50 network to extract feature pyramids from the 1st and 2nd input frames. 
These feature pyramids have 4 levels with a coarsest scale of 1/32 of the original resolution.
For flow input, we first use conv7$\times$7 and conv5$\times$5 layers with stride 2 to get features at 1/4 resolution.
Then we apply 4 blocks of 2 conv3$\times$3 layers to get pyramid flow features with 4 scales.
We also apply residual blocks identical to~\cite{xian2018monocular} to project each image feature map to 256-channels, and each flow feature map to 128-channels. Finally, we warp the features in the pyramid for the 2nd frame to the 1st using the flow.

At each level of the feature pyramids, we concatenate the feature map of the 1st frame and the warped feature of the 2nd frame, and then use a residual block to project it to 256-channels, and concatenated it with the flow features.
A final residual block projects this tensor to 256-channels.

\vspace{5px} \noindent \textbf{Depth decoder}
The fused feature pyramid is fed into a depth decoder with skip connections. Starting from the coarsest scale, the output from previous scale of the decoder is bilinearly upsampled and then added to the corresponding fused feature from the pyramid. 
After reaching 1/4 of the full resolution, a stack of 2 $3\times3$ conv layers and 1 bilinear upsampling layer is applied to produce final full-resolution depth prediction in log scale.

As shown in Figure~\ref{fig:ablation} and our ablation study, we find that compared to using features only from single image, fusing features from the second frame and the flow helps identify foreground objects and produces more accurate result. 



%% file: sections/evaluation.tex
\begin{figure*}[t]
    \centering
    \begin{tabular}{*{4}{c@{\hspace{2px}}}}
    \includegraphics[width=.2\linewidth]{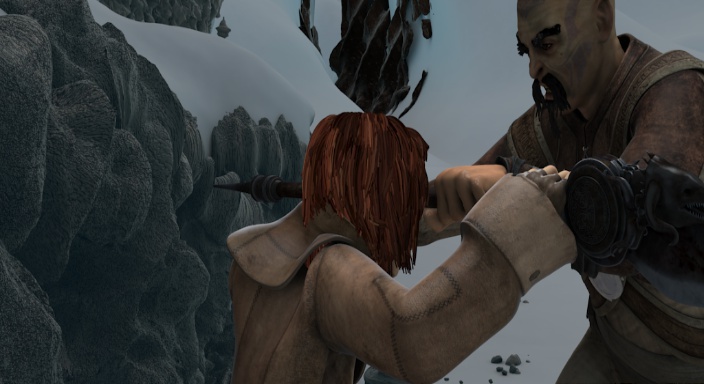}&
    \includegraphics[width=.2\linewidth]{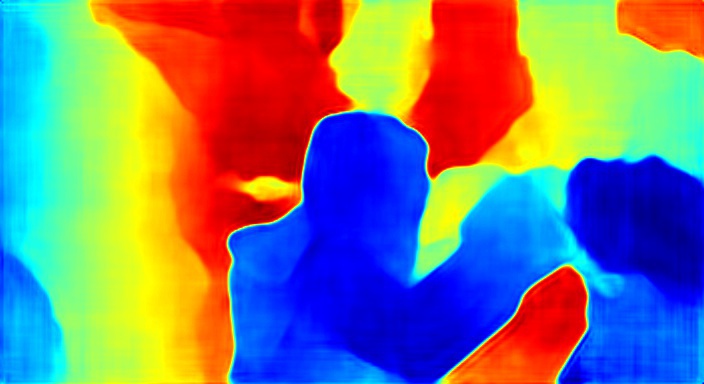}&
    \includegraphics[width=.2\linewidth]{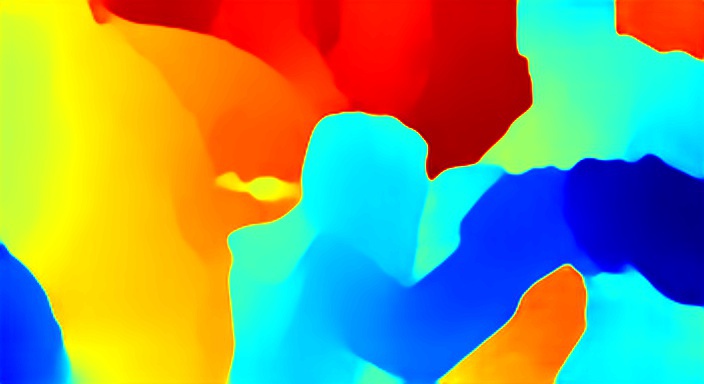}&
    \includegraphics[width=.2\linewidth]{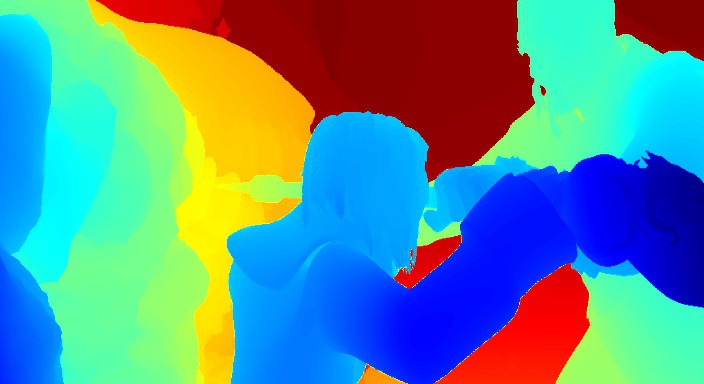}\\
    \includegraphics[width=.2\linewidth]{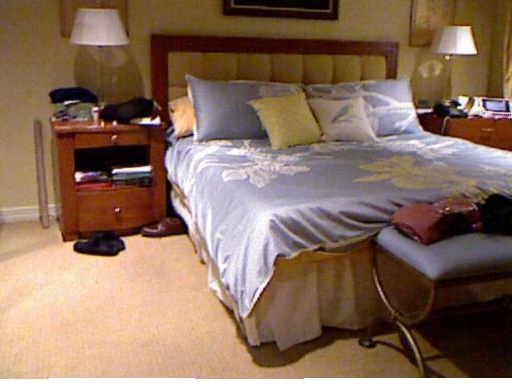}&
    \includegraphics[width=.2\linewidth]{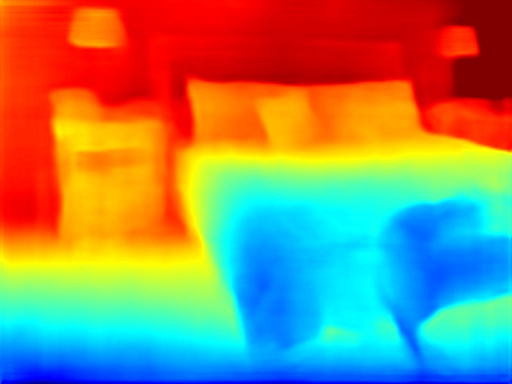}&
    \includegraphics[width=.2\linewidth]{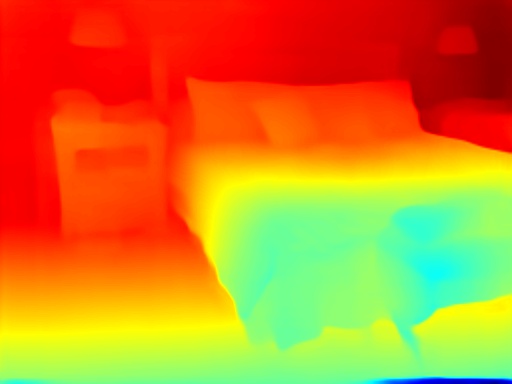}&
    \includegraphics[width=.2\linewidth]{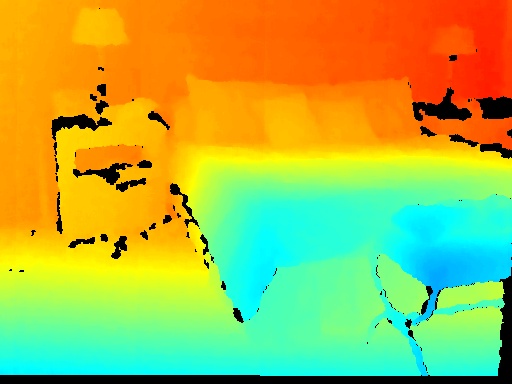}\\
    Input & Ordinal loss~\cite{xian2018monocular} & NMG (Ours) & GT \\
    \end{tabular}
    \caption{Comparison of NMG with an Ordinal loss. Top: trained on \dataset{} and tested on SINTEL; Bottom: trained and tested on NYU-v2.}
    \label{fig:ordinal}
\end{figure*}

\section{Evaluation}
\label{sec:eval}
We evaluate our network on three video datasets with non-rigid scenes, and compare to other methods trained on a variety of training sets as well as a traditional geometric method (COLMAP~\cite{schoenberger2016mvs}).
We seek to answer the following questions:
\begin{itemize}[topsep=0pt,itemsep=-1ex,partopsep=1ex,parsep=1ex]
    \item How does the proposed NMG loss compared to ordinal loss used by previous methods?
    \item How important is the temporal information used by our network to improve depth prediction?
    \item How does \dataset{} compare to other multi-view datasets when used as training source?
    \item How does our method generalize to other dataset compared to state-of-the-art methods?
\end{itemize}

\vspace{5px} \noindent\textbf{Experiment setup}
We hold out 50 videos, and sample 597 frame pairs as our testing set. 
We use all the rest videos in our dataset for training and validation. 
The validation set consists of 1000 frames from 500 randomly sampled clips. 
To avoid bias in sampling from longer videos, we randomly sample 1 clip per video at each training epoch. 
We train our network using Adam with default parameters and use batchsize of 6.
For our ablation study, we also train a single-view depth prediction network which has identical architecture of our proposed network, but without the feature pyramids from the flow and the second input frame. 

\vspace{5px} \noindent\textbf{Metrics}
For comparison on SINTEL, we use the commonly used mean relative error (MRE) and scale invariant logarithmic error (SILog) evaluated on inverse depth.
These two errors are measured for pixels up to 20 meters away.
We also use the proposed NMG as an additional metric.
To avoid bias towards testing samples with large inverse depth value, we scaled the NMG error by the mean of the ground truth inverse depth.

For KITTI, we use the 697 test samples (parired with consecutive frame for multi-view testing) from the Eigen split. We report absolute relative difference (abs rel), squared relative difference (sqr rel), relative mean squared logarithmic error (RMSE log) and percentage of error lower than a threshold (e.g. $\delta<1.25$).
Since our method cannot predict metric depth, we align the prediction to the groundtruth by the median of the depth map, as is also done in ~\cite{zhou2017unsupervised}.

In addition, we perform evalution on our \dataset{} test set. We compute the NMG error excluding pixels which fail the left-right disparity consistency check (visualized as black region in Figure~\ref{fig:ordinal}). While we do not claim that this is any indication of the generalization ability of our approach, we do so as an indication of how methods might perform on diverse non-rigid scenes.

\subsection{Evaluation of NMG Loss}
In order to evaluate the effect of our normalized multiscale gradient (NMG) loss, we compare it to the ordinal loss used in RedWeb~\cite{xian2018monocular}.
To do this, we first compare both losses on the NYU-v2 dataset~\cite{nyu_v2}. To mimic disparity maps from Internet stereoscopic images, we apply affine transformations to the ground truth depth values with uniformly sampled slope and bias parameters. These synthesized disparity maps are then used to train a single view depth estimator with different loss functions. As shown in Table~\ref{tab:ordinal_nyu}, NMG loss has lower testing errors compared to ordinal loss, and its depth map prediction appears to be smoother as well (see Fig.~\ref{fig:ordinal}).

Moreover, we train the proposed multi-view depth estimator 
 on \dataset{} with both losses. We see in Table~\ref{tab:ordinal}, and Figure~\ref{fig:ordinal}, that our loss function again yields visually smoother, and more accurate depth maps when tested on SINTEL.

\begin{table}[ht]
\centering
\rowcolors{2}{}{rowblue}
\begin{small}
\begin{tabular}{l||c|cc}
\toprule
 & loss by Eigen~\cite{eigen2014depth} & ordinal & NMG\\
 \hline
 \hline
 train label & depth & \{synthesized & disparity\}\\
 \hline
 rmse &  0.467 & 0.767 & \textbf{0.706}\\ 
 abs rel & 0.128 & 0.184 & \textbf{0.164}\\ 
 $\delta<1.25$ & 0.840 & 0.753 & \textbf{0.768}\\
 $\delta<1.25^2$ & 0.961 & 0.927 & \textbf{0.945}\\
 $\delta<1.25^3$ & 0.990 & 0.0.979 & \textbf{0.988}\\
\bottomrule
\end{tabular}
\end{small}
\caption{Comparison between NMG and ordinal loss on NYU-v2. Trained using synthesized disparity map with randomly sampled camera parameters. Result of directly using depth map as training label (loss by Eigen~\cite{eigen2014depth}) is also provided as reference.}
\label{tab:ordinal_nyu}
\end{table}

\begin{table}[ht]
\centering
\rowcolors{2}{}{rowblue}
\begin{tabular}{l|ccc}
\toprule
     & NMG & MRE & SILog\\
\hline
    ordinal ~\cite{xian2018monocular} & 0.963 & 0.350 & 0.228\\
    NMG (Ours) & \textbf{0.890} & \textbf{0.311} & \textbf{0.172}\\
\bottomrule
\end{tabular}
\caption{Comparison between NMG and ordinal loss. Trained on \dataset{} and tested on SINTEL. Columns show different evaluation metrics.}
\label{tab:ordinal}
\end{table}



\begin{figure}[t]
    \centering
    \begin{tabular}{*{3}{c@{\hspace{3px}}}}
    \includegraphics[width=.3\linewidth]{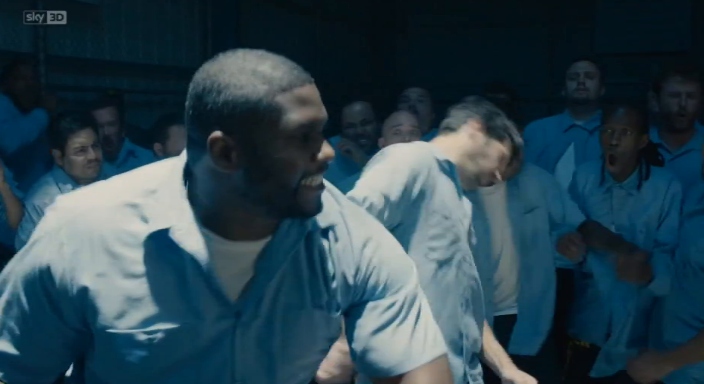} & 
    \includegraphics[width=.3\linewidth]{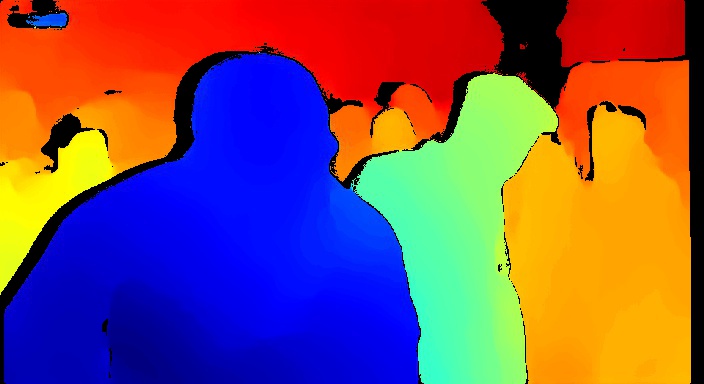} &
    \includegraphics[width=.3\linewidth]{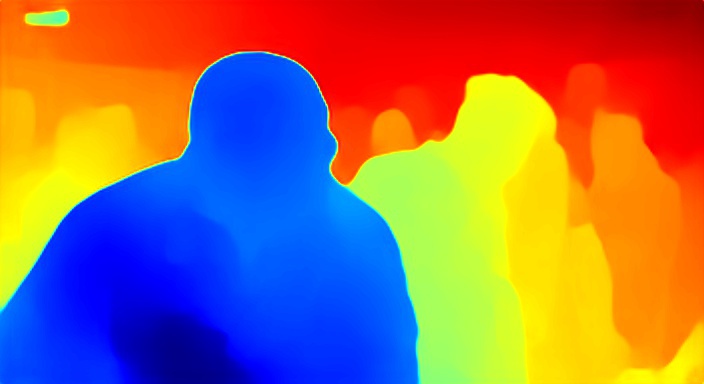}\\
    \small{Input} & \small{GT} & \small{Ours}\\
    \includegraphics[width=.3\linewidth]{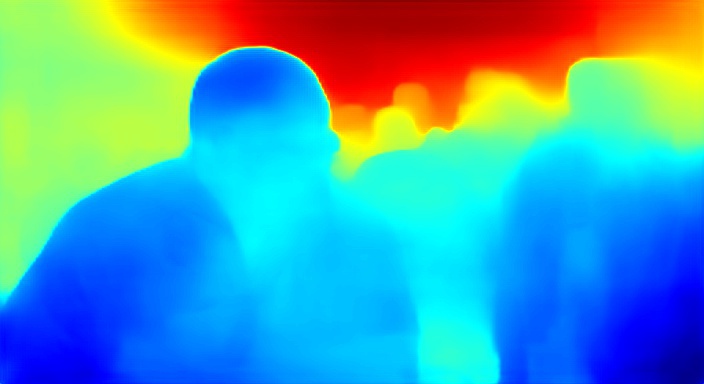} &
    \includegraphics[width=.3\linewidth]{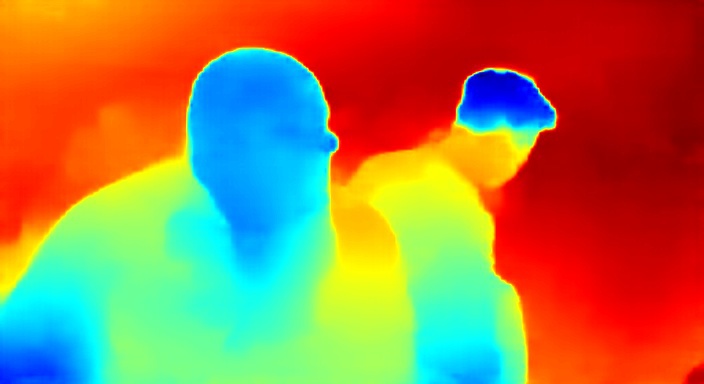} &
    \includegraphics[width=.3\linewidth]{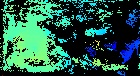}\\
     \small{MegaDepth} & \small{DEMON} & \small{COLMAP}\\
    \end{tabular}
    \caption{Qualitative comparison between our multi-view depth network and the state-of-the-art single/multi-view depth prediction methods.}
    \label{fig:comp_sota}
\end{figure}

\begin{figure*}[t]
    \centering
    \begin{tabular}{*{5}{c@{\hspace{2px}}}}
    \includegraphics[width=.19\linewidth]{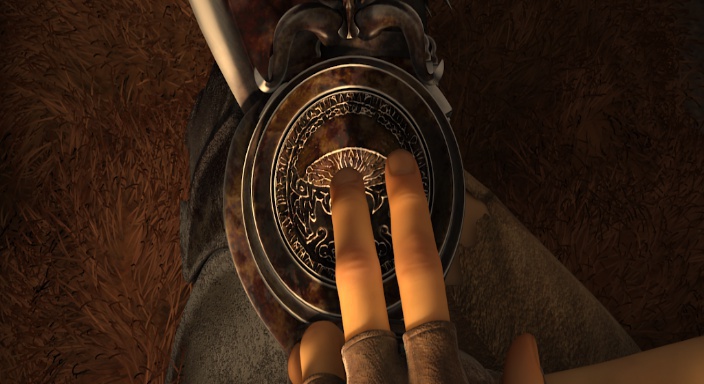} &
    \includegraphics[width=.19\linewidth]{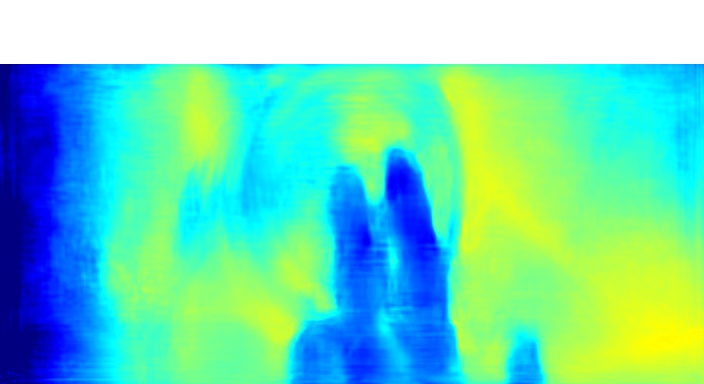}&
    \includegraphics[width=.19\linewidth]{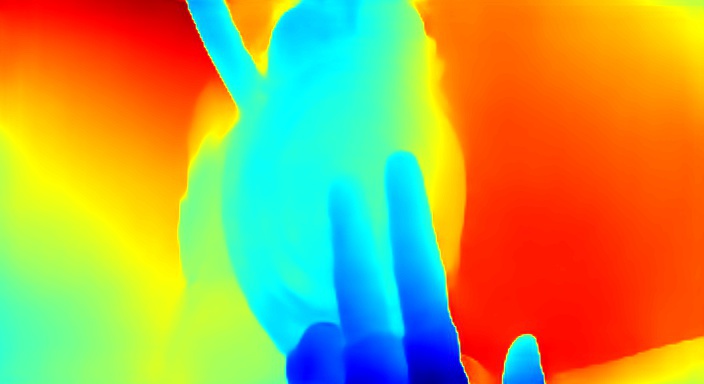}&
    \includegraphics[width=.19\linewidth]{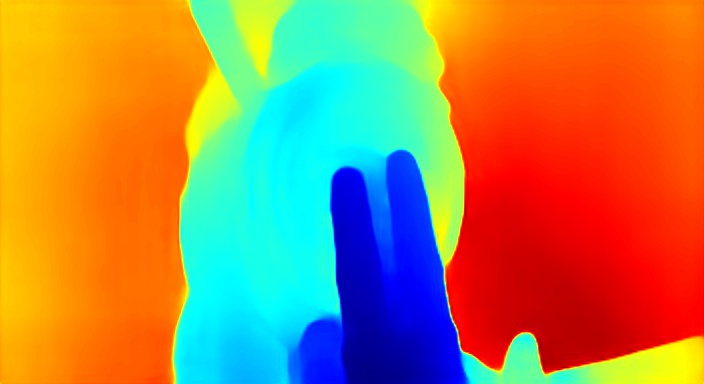}&
    \includegraphics[width=.19\linewidth]{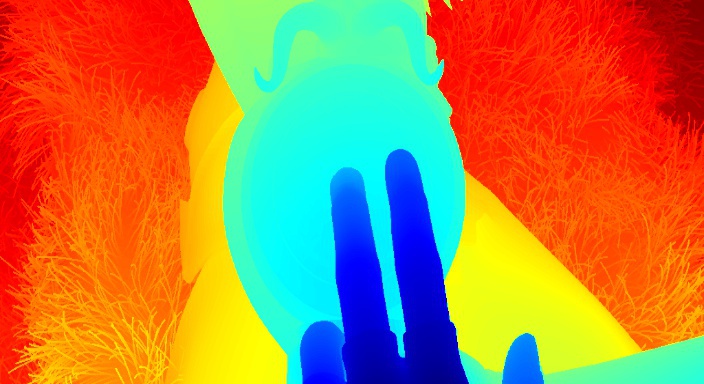}\\
    \includegraphics[width=.19\linewidth]{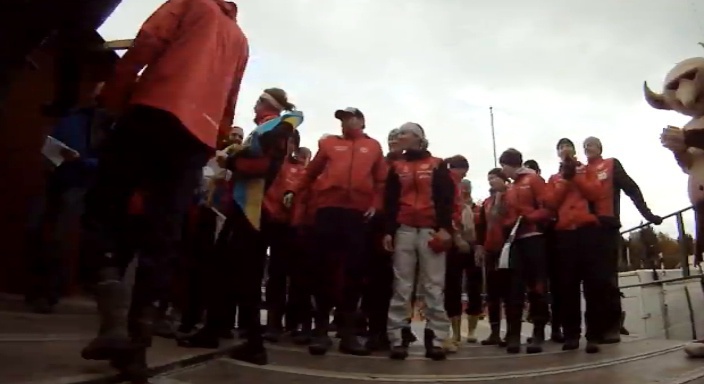}&
    \includegraphics[width=.19\linewidth]{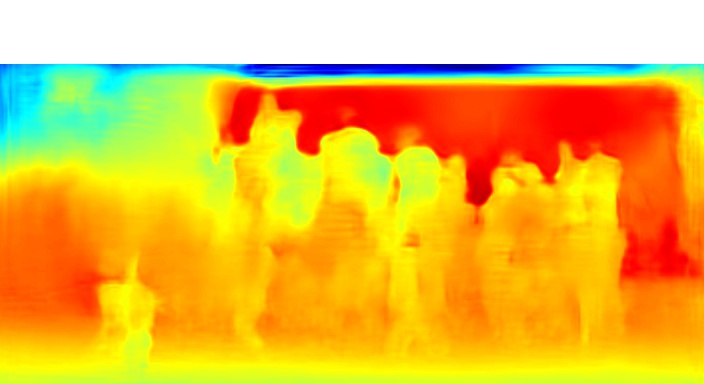}&
    \includegraphics[width=.19\linewidth]{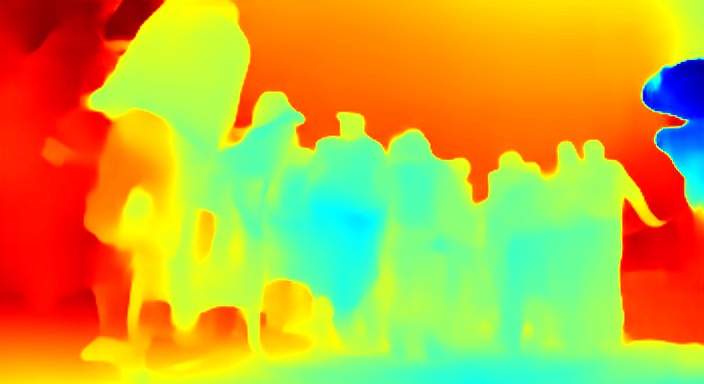}&
    \includegraphics[width=.19\linewidth]{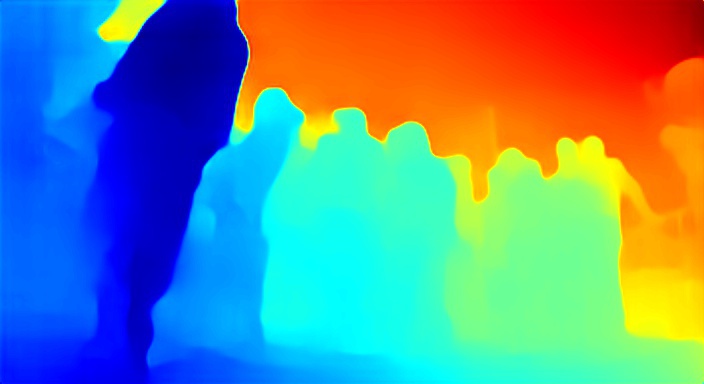}&
    \includegraphics[width=.19\linewidth]{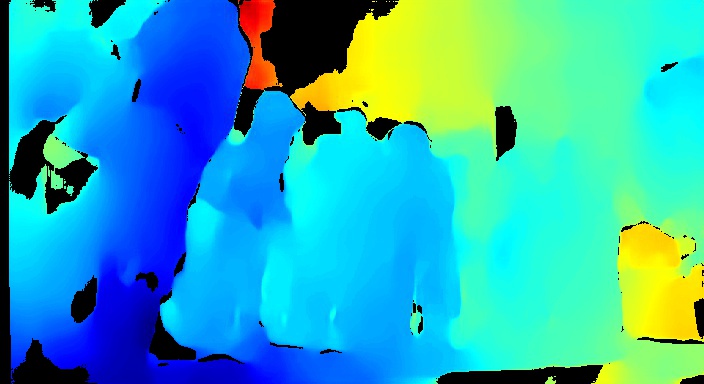}\\
    Input & KITTI  & DEMON & \dataset{} (Ours) & GT \\
    \end{tabular}
    \caption{We show the performance of the same multi-view depth network trained on different datasets, and tested on Sintel and \dataset{}. Note that KITTI does not have groundtruth for the top region of the image, thus network trained with KITTI may produce arbitrary values in this area. For fairer comparison, we cropped out the top part.}
    \label{fig:datasets}
\end{figure*}

\begin{table*}[]
\rowcolors{2}{}{rowblue}

\resizebox{\linewidth}{!}{
\begin{footnotesize}
\begin{tabular}{l|l|cccc|ccc}
\toprule
 & & & single view &  & & & multi view &\\
 \hline
 Test set & Metric & MegaDepth~\cite{li2018megadepth} & ReDWeb$^*$ & DDVO~\cite{Wang_2018_CVPR} & Ours & COLMAP & DEMON~\cite{ummenhofer2017demon} & Ours\\ 
\hline
SINTEL  & MRE    & 0.364 & 0.401 & 0.435 & 0.333 & - & 0.478   &  \textbf{0.311}\\
        & SILog  & 0.266 & 0.311 & 0.360 &  0.206 & - & 0.397  & \textbf{0.172}\\
& NMG   & 1.212 & 1.041 & 1.492 & 0.993 & - & 1.311 & \textbf{0.890}\\
\hline
KITTI & abs rel       & 0.220 & 0.234 & \textbf{0.148} & 0.230 & - & 0.235  & \underline{0.213}\\ 
      & $\delta<1.25$ & 0.632 & 0.617 & \textbf{0.812} & 0.606 & - & 0.605 & \underline{0.637}\\
\hline
KITTI      & abs rel       & 0.227 & 0.231 & \textbf{0.183} & 0.230 & - & 0.307 & \underline{0.207}\\
   pedestrian        & $\delta<1.25$ & 0.625 & 0.622 & \textbf{0.746} &  0.610 & - & 0.458  & \underline{0.654}\\
\hline
\hline
\dataset{} & NMG & 1.269 & 1.152 & 1.505 &  1.012 & 2.021 (69.8\%) & 1.418 & \textbf{0.899} \\
\bottomrule
\end{tabular} 
\end{footnotesize}
}
\caption{Quantitative comparison of methods. All metrics in this table are \emph{lower is better} except $\delta<1.25$. Best results are shown in bold, and underlining indicates the best result second only to DDVO trained on KITTI. COLMAP returns semi-dense depth values. In this case, 69.8\% of pixels had valid depths computed, we compute the metric over these pixels only. $^*$ numbers reported for RedWeb is from our re-implementation of ~\cite{xian2018monocular} due to their code has not been released.}
\label{tab:quantitative}

\end{table*}

\subsection{Multi-view v.s. Single-view Prediction}

\begin{figure*}[t]
    \centering
    \begin{tabular}{*{5}{c@{\hspace{2px}}}}
    \begin{tikzpicture}[inner sep=0]
        \node (fig1) at (0,0)
        {\includegraphics[width=.18\linewidth]{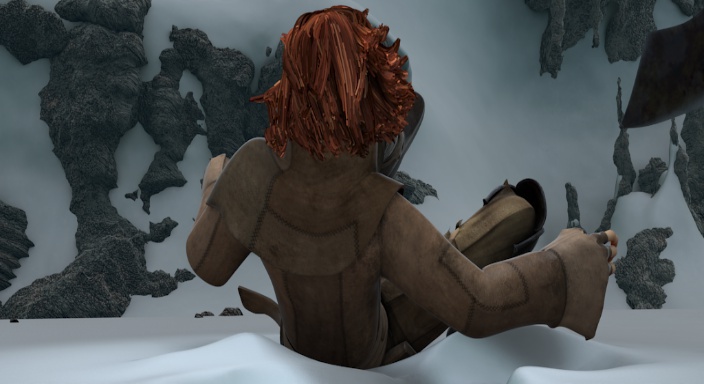}};
        \node (fig2) at (0.87,-.46)
        {\includegraphics[width=0.07\linewidth]{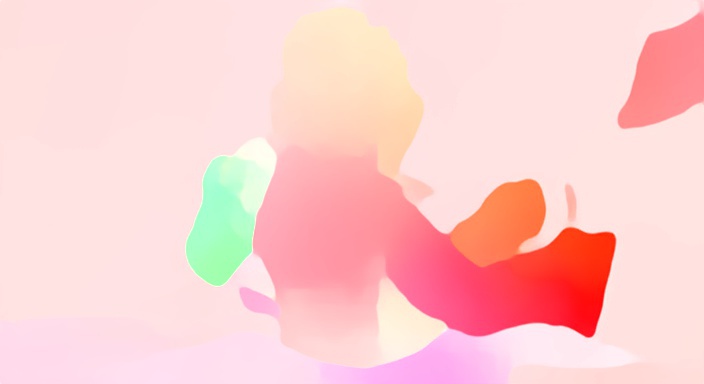}};  
    \end{tikzpicture}&
    \includegraphics[width=.18\linewidth]{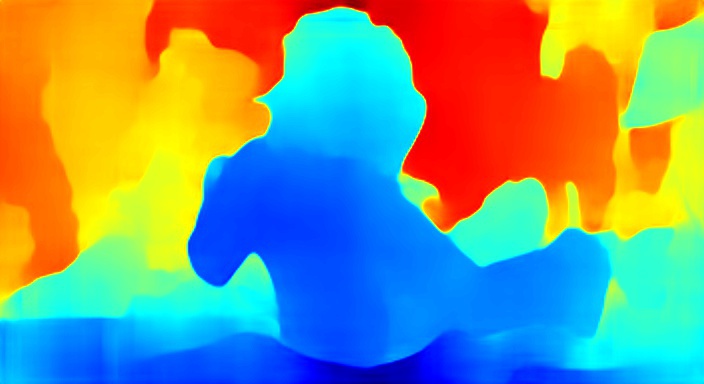}&
    \includegraphics[width=.18\linewidth]{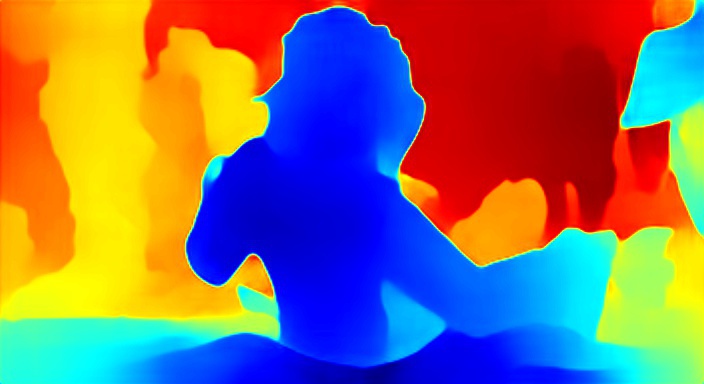}&
    \includegraphics[width=.18\linewidth]{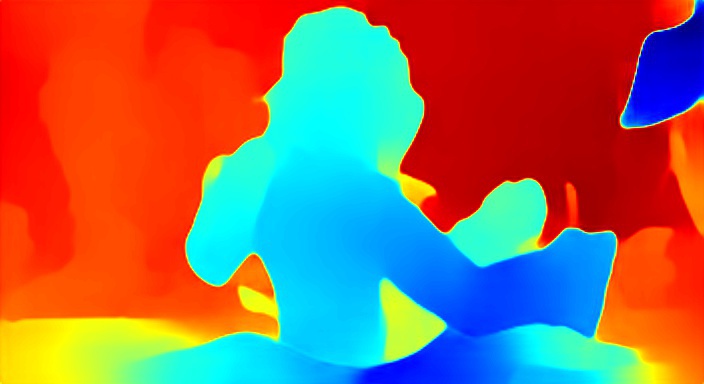}&
    \includegraphics[width=.18\linewidth]{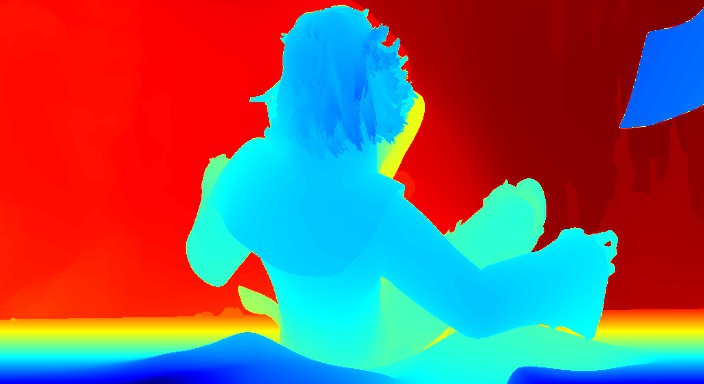}\\
    \begin{tikzpicture}[inner sep=0]
        \node (fig1) at (0,0)
        {\includegraphics[width=.18\linewidth]{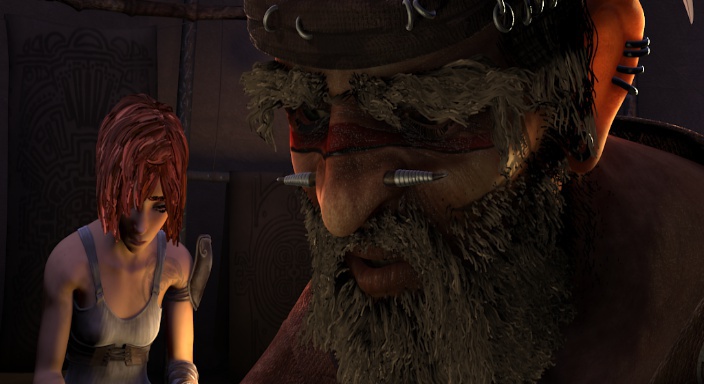}};
        \node (fig2) at (0.87,-.46)
        {\includegraphics[width=0.07\linewidth]{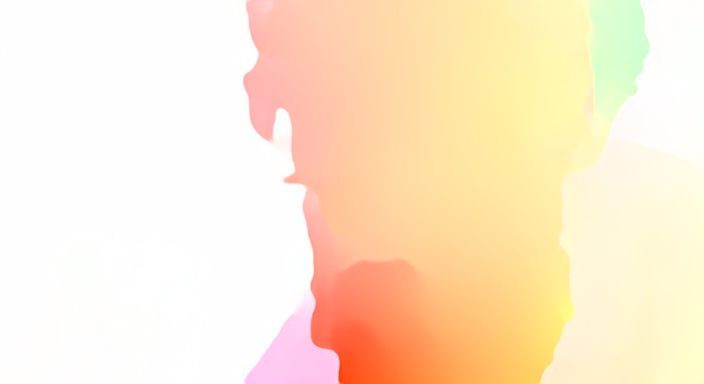}};  
    \end{tikzpicture}&
    \includegraphics[width=.18\linewidth]{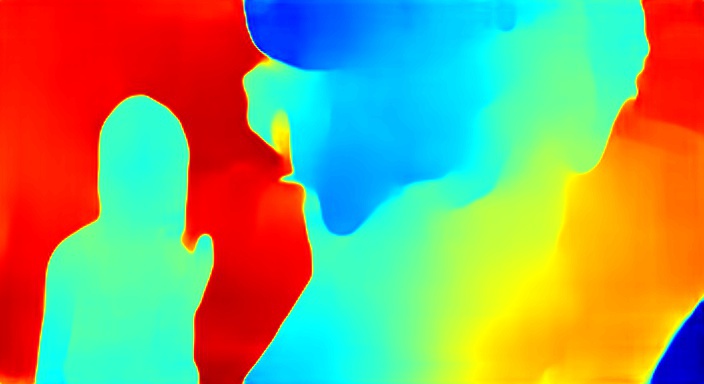}&
    \includegraphics[width=.18\linewidth]{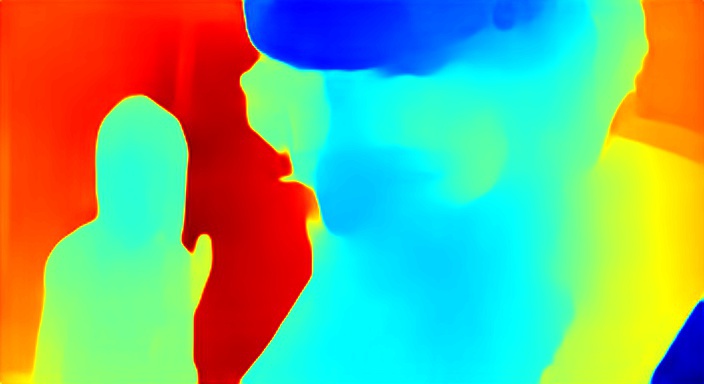}&
    \includegraphics[width=.18\linewidth]{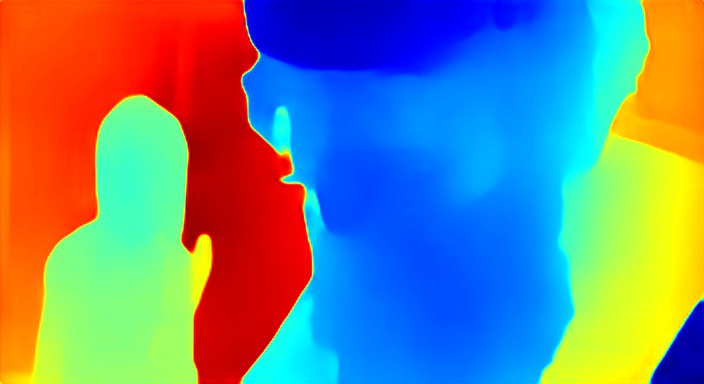}&
    \includegraphics[width=.18\linewidth]{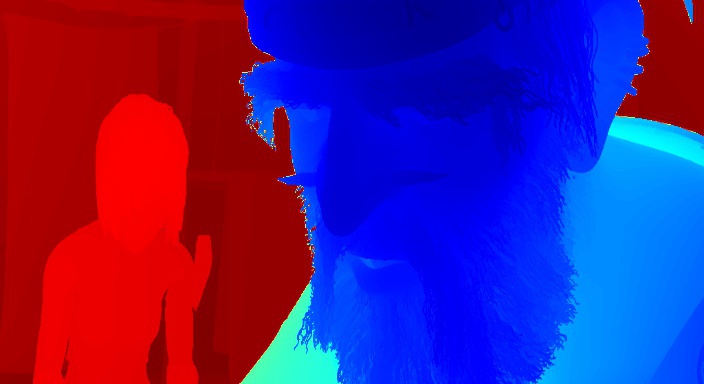}\\
    \begin{tikzpicture}[inner sep=0]
        \node (fig1) at (0,0)
        {\includegraphics[width=.18\linewidth]{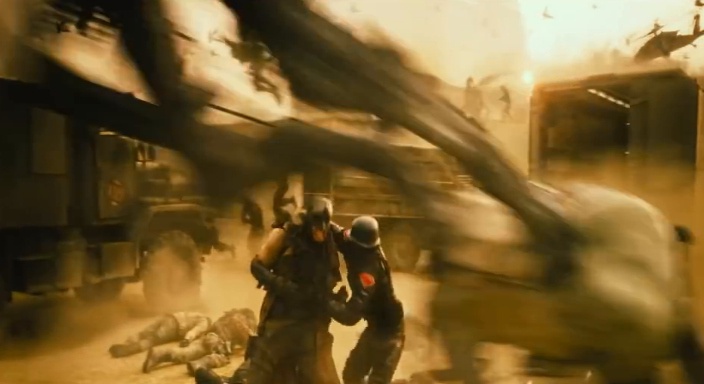}};
        \node (fig2) at (0.87,-.46)
        {\includegraphics[width=0.07\linewidth]{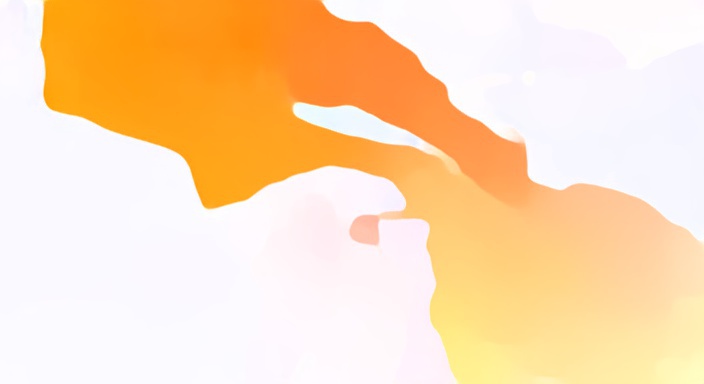}};  
    \end{tikzpicture}&
    \includegraphics[width=.18\linewidth]{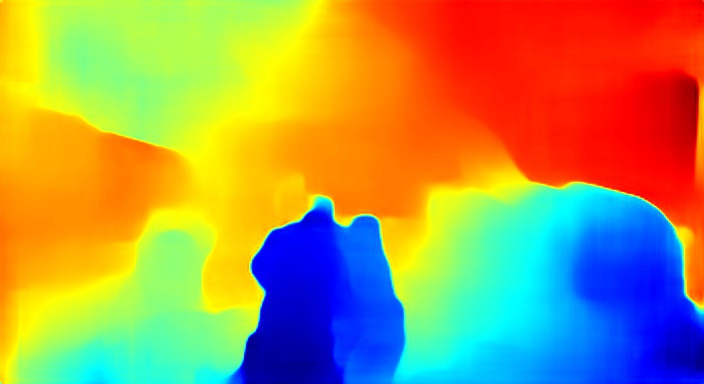}&
    \includegraphics[width=.18\linewidth]{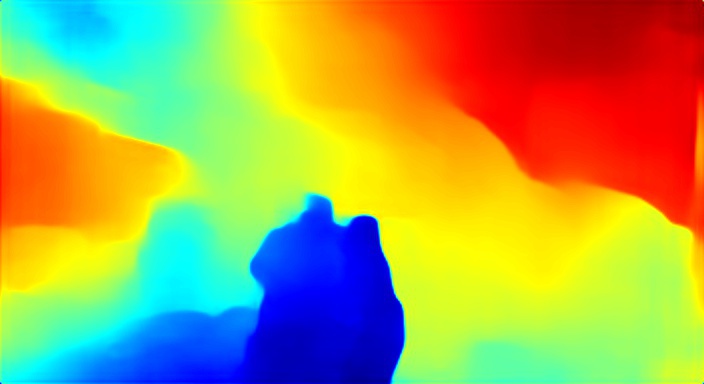}&
    \includegraphics[width=.18\linewidth]{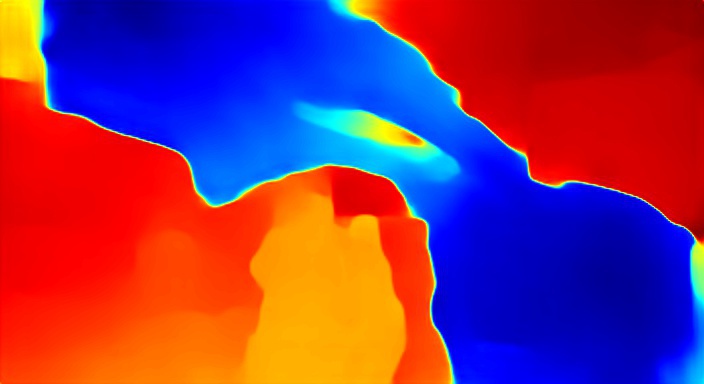}&
    \includegraphics[width=.18\linewidth]{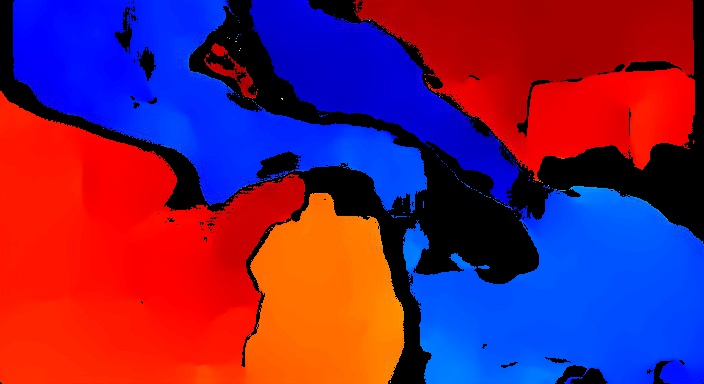}\\
    Input & $I^t$ & $I^t$, $I^t$ + 0 flow & $I^t$, $I^{t+1}$ + flow& GT \\
    \end{tabular}
    \caption{Depth maps predicted with different inputs, showing the impact of additional temporal information made available to the network. Using sequential frames and optical flow (4th column) improves baselines without temporal information.}
    \label{fig:ablation}
\end{figure*}

We evaluate the effect of incorporating temporal information at \emph{test} time, by comparing our approach against two baselines: 1) a single view depth prediction network; 2) our proposed network with two identical frames ($I_t$, $I_t$) and a zero flow map as input. This baseline is used to verify if the improvement is truly from temporal information, instead of the difference in network architecture.
Table~\ref{tab:ablation}, and Figure~\ref{fig:ablation} show that the two baselines achieve similar performance, our proposed method with additional temporal information as input can identify foreground objects better and gives more accurate depth estimation.
\begin{table}[ht]
   \centering
    \rowcolors{4}{}{rowblue}
    \resizebox{.75\linewidth}{!}{
    \begin{tabular}{l|ccc}
    \toprule
        \small{Training set}  & & \small{Multi-view test set} &\\
        \hline
        & SINTEL  & KITTI & \dataset{} \\
        & \footnotesize{SILog} & \footnotesize{RSME(log)}& \footnotesize{NMG}\\
        \hline
        KITTI & 0.3871 & \textbf{0.180} & 1.620\\
        DEMON & 0.222  & 0.356 & 1.205\\
        \dataset{} & \textbf{0.172} &  0.317 & \textbf{0.899}\\
        \bottomrule
    \end{tabular}
    }
    \caption{Testing result of our proposed network trained on different datasets. DEMON refers to the video datasets used for training in~\cite{ummenhofer2017demon}, SUN3D, RGBD, and Scenes11.}
    \label{tab:dataset_cross_val}
\end{table}

\subsection{Training on Different Multi-view Datasets}
We conduct a controlled experiment of training our multi-view depth prediction network on different multi-view datasets and compare their cross dataset generalization performance. We use scale invariant logarithm loss~\cite{eigen2014depth} to train on KITTI; and a combination of scale invariant logarithm depth loss and gradient loss to train on the training set (SUN3D+ RGBD + Scene11) proposed by DEMON~\cite{ummenhofer2017demon}, consisting mostly of rigid scenes.  As shown in Table~\ref{tab:dataset_cross_val} and Figure~\ref{fig:datasets}, training on our \dataset{} dataset has the lowest error on Sintel and outperforms the DEMON training set on KITTI, while models learned from KITTI have the worst performance on other datasets.

\begin{table}[ht]
    \centering
    \resizebox{\linewidth}{!}{
    \rowcolors{2}{}{rowblue}
    \begin{tabular}{l|ccc|ccc|c}
    \toprule
        & & SINTEL &  &  & KITTI & & \dataset{}\\
        Input &  NMG & MRE & SILog & abs rel & sq rel & RMSE(log) & NMG \\
        \midrule
        $I^t$ & 0.993 & 0.333 & 0.206 & 0.230 & 1.937 & 0.327 & 1.012\\
        $I^t,I^{t}$, 0 flow & 0.972 & 0.326 & 0.198 & 0.235 & 2.103 & 0.341 & 0.977\\
        $I^t,I^{t+1}, $flow  & \textbf{0.890} & \textbf{0.311} & \textbf{0.172} & \textbf{0.213} & \textbf{1.849} & \textbf{0.317} & \textbf{0.899}\\
    \bottomrule
    \end{tabular}
    }
    \caption{Evaluate the improvement due to adding the flow and second frames as input in test time. Each row describes the input to the network. i.e.  ($I^t,I^{t}$) means the same image is given twice; ``0-flow''means feeding flow map filled with zeros.}
    \label{tab:ablation}
\end{table}

\subsection{Cross Dataset Evaluation}
We compare the generalization capability of the comapred methods by testing on different video datasets, i.e. SINTEL, KITTI and \dataset{}. We notice that most of the scenes in the KITTI test set are rigid, therefore, we select a subset of 24 images with pedestrians, and use it to analyze the performance of handling non-rigid scenes for the compared methods.

Table~\ref{tab:quantitative} and Figure~\ref{fig:comp_sota} show that our (single/multi-view) network compares favorably to single view depth prediction methods, i.e. MegaDepth~\cite{li2018megadepth}, our re-implementation of RedWeb~\cite{xian2018monocular} and DDVO~\cite{Wang_2018_CVPR}, which is unsupervised learned on KITTI .  
On the KITTI test set, MegaDepth performs similarly to ours, which could be due to their abundant training for street and landmark scenes, while our method demonstrates better result on the pedestrian subset. Not surprisingly, DDVO performs the best on KTTI since it is trained on this dataset, but it generalizes poorly on other test sets.

We also run DEMON~\cite{ummenhofer2017demon} -- a deep learning method for predicting depth and camera pose from two views. We find that DEMON produces plausible results for rigid scenes with sufficient parallax, but generates worse results on nonrigid scenes.  In addition, we compare the quality of reconstruction on some clips from our dataset using COLMAP~\cite{schonberger2016structure}, and show quantitative results on Table~\ref{tab:quantitative}. 
We note that due to running time constraints we ran COLMAP on a random subset of our data.

Throughout the evaluation above, our method consistently outperforms our single-view depth prediction baseline, which indicates that our performance gain is not solely due to our training set but also from the proposed network's ability to take temporal information into account.

Finally, geometric-based methods by Kumar et al.~\cite{kumar2017monocular} and Ranftl et al.~\cite{dmde} are related to ours, but we're unable to provide meaningful comparison here due to their code and result is not publicly available, and the numbers reported in their paper are from an undisclosed subset of the dataset.

%% file: sections/discussion.tex
\begin{figure}
    \centering
    \begin{tabular}{*{3}{c@{\hspace{3px}}}}
    \includegraphics[width=.31\linewidth]{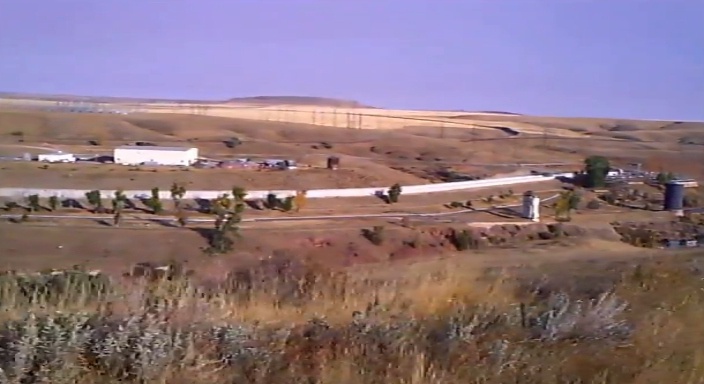} &
    \includegraphics[width=.31\linewidth]{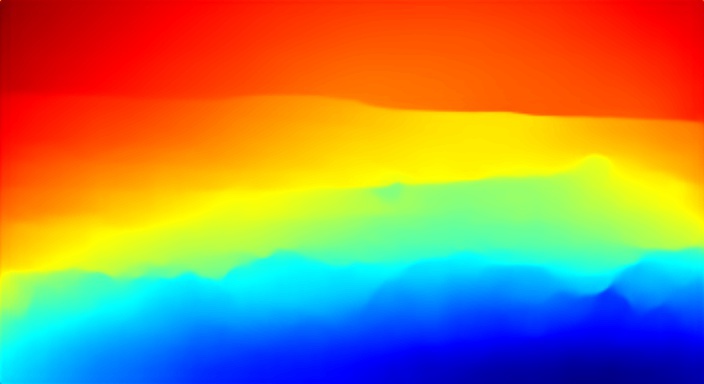} &
    \includegraphics[width=.31\linewidth]{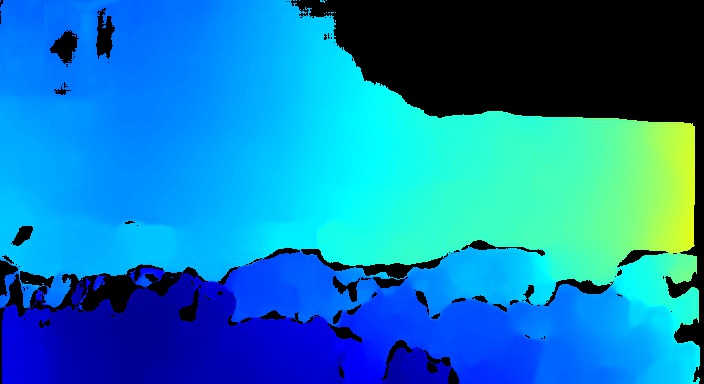} \\
    Input & Ours & GT from stereo \\
    \end{tabular}
    \caption{Limitations. Stereo supervision is less reliable at long distances or texture-less regions.}
    \label{fig:limitations}
\end{figure}

\subsection{Limitations}
One limitation of our approach, is that by using stereo disparity as supervision, we are restricting ourselves to scenes whose disparity can be computed. 
This can be problematic with large-scale scenes such as landscapes, where stereo baselines would have to be very large to have nonzero disparity.
In Fig~\ref{fig:limitations}, we can see that our ``ground-truth'' is in fact incorrect in such scenes. 
We can also have similar difficulties in untextured regions, where reliable correspondences do not exist for supervision.
In general, any biases in the reconstruction step will likely persist in our final results.

\section{Conclusion}
In conclusion, we present a step towards data-driven reconstruction of non-rigid scenes, by introducing the first in-the-wild stereo video dataset that features a wide distribution of nonrigid object types.
We hope that this dataset will encourage future work in the area of diverse non-rigid video reconstruction, a topic with many exciting applications.

%% file: sections/appendix.tex
\begin{figure}[t]
    \centering
    \includegraphics[width=\linewidth]{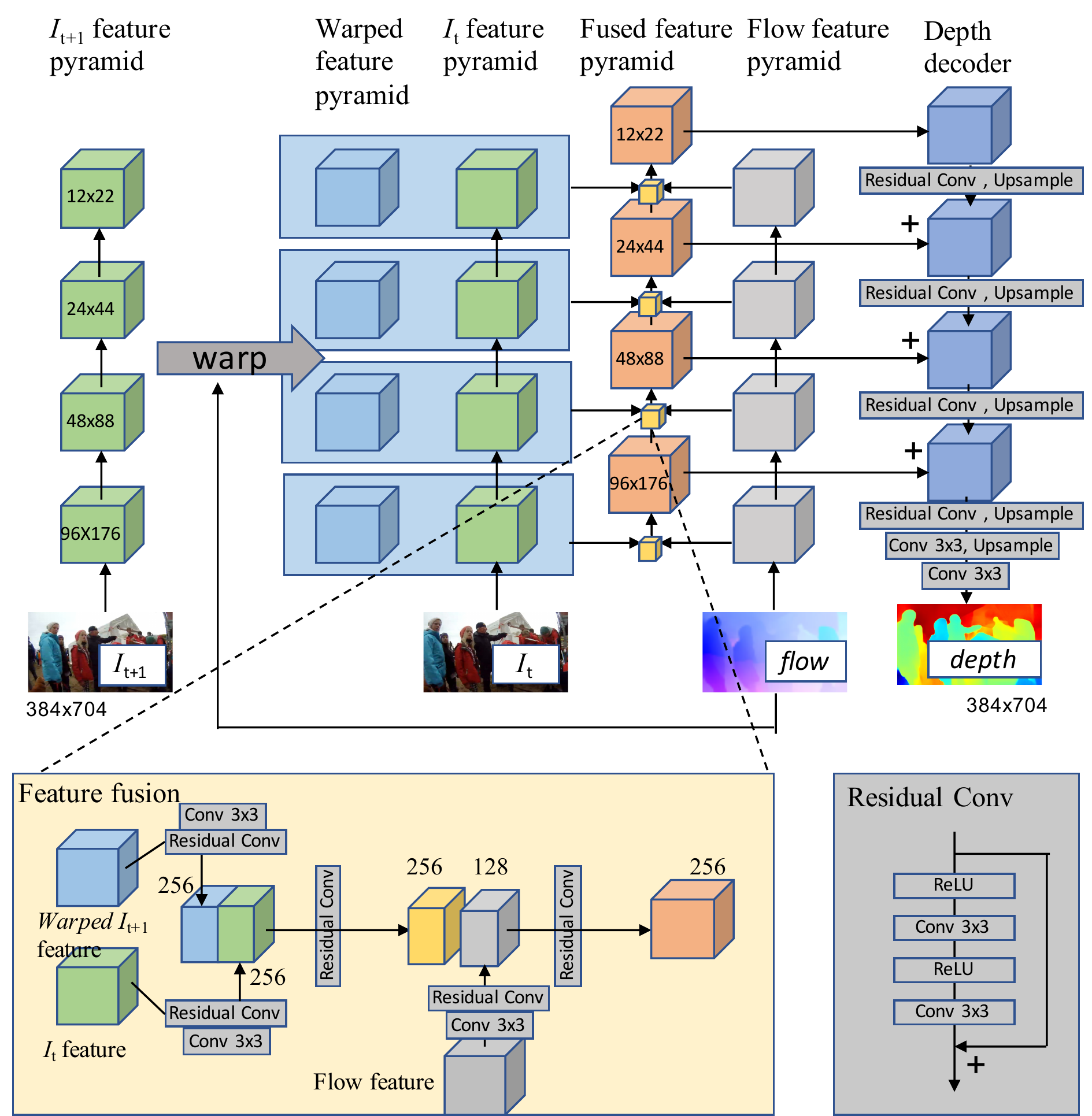}
    \caption{Our architecture consists of three feature extraction pyramids. The features extracted from the pyramid for frame $I_t$ is concatenated with features extracted from the next frame $I_{t+1}$, warped into $I_t$ using the flow computed between the two. This is fused with the feature pyramid extracted from the flow, and is then decoded into the final depth (right column).
    Feature fusion is shown in detail below (bottom left), it consists of concatenation and reprojection operations using residual convolution blocks (bottom right). 
    }
    \label{fig:network}
\end{figure}

\section{Network architecture}
\label{appendix:architecture}
Figure~\ref{fig:network} shows our two-frame network architecture for depth prediction. Please see the caption for more detail.

\section{Additional samples from \dataset{}}
We include additional samples from our \dataset{} dataset in Figure~\ref{fig:supp_dataset}. Each sample is rendered with an RGB frame, estimated disparity map, and optical flow.

\section{Additional qualitative results}
Additional qualitative comparison on \dataset{} test set is included in Figure~\ref{fig:supp_yt3d_comp}. We compare our method, which takes sequential frames $I_t$, $I_{t+1}$ and an optical flow map as input, to two baseline monocular (single-frame) depth predictors: 1) the hourglass network provided by MegaDepth~\cite{li2018megadepth}; 2) our network architecture, but with the feature streams from optical flow and next frame removed. All architectures are trained with our ~\dataset{} dataset using the same loss function. As shown in the figure, our method produces more accurate estimation especially in the foreground regions.

We also include qualitative results on KITTI test set in Figure~\ref{fig:supp_kitti_comp}. Compared to the baselines, our method gives more details in pedestrians and light poles. 




\begin{figure*}
\centering
\begin{tabular}{*{9}{c@{\hspace{2px}}}}
Image & Disparity & Flow & Image & Disparity & Flow & Image & Disparity & Flow \\
\includegraphics[width=.11\linewidth]{figures/supp_dataset/0004552.jpg} &
\includegraphics[width=.11\linewidth]{figures/supp_dataset/0004552_d.jpg} &
\includegraphics[width=.11\linewidth]{figures/supp_dataset/0004552_flo.jpg} &
\includegraphics[width=.11\linewidth]{figures/supp_dataset/0009944.jpg} &
\includegraphics[width=.11\linewidth]{figures/supp_dataset/0009944_d.jpg} &
\includegraphics[width=.11\linewidth]{figures/supp_dataset/0009944_flo.jpg} &
\includegraphics[width=.11\linewidth]{figures/supp_dataset/0012865.jpg} &
\includegraphics[width=.11\linewidth]{figures/supp_dataset/0012865_d.jpg} &
\includegraphics[width=.11\linewidth]{figures/supp_dataset/0012865_flo.jpg} \\ 
\includegraphics[width=.11\linewidth]{figures/supp_dataset/0001539.jpg} &
\includegraphics[width=.11\linewidth]{figures/supp_dataset/0001539_d.jpg} &
\includegraphics[width=.11\linewidth]{figures/supp_dataset/0001539_flo.jpg} &
\includegraphics[width=.11\linewidth]{figures/supp_dataset/0001390.jpg} &
\includegraphics[width=.11\linewidth]{figures/supp_dataset/0001390_d.jpg} &
\includegraphics[width=.11\linewidth]{figures/supp_dataset/0001390_flo.jpg} &
\includegraphics[width=.11\linewidth]{figures/supp_dataset/0003589.jpg} &
\includegraphics[width=.11\linewidth]{figures/supp_dataset/0003589_d.jpg} &
\includegraphics[width=.11\linewidth]{figures/supp_dataset/0003589_flo.jpg} \\ 
\includegraphics[width=.11\linewidth]{figures/supp_dataset/0001359.jpg} &
\includegraphics[width=.11\linewidth]{figures/supp_dataset/0001359_d.jpg} &
\includegraphics[width=.11\linewidth]{figures/supp_dataset/0001359_flo.jpg} &
\includegraphics[width=.11\linewidth]{figures/supp_dataset/0007079.jpg} &
\includegraphics[width=.11\linewidth]{figures/supp_dataset/0007079_d.jpg} &
\includegraphics[width=.11\linewidth]{figures/supp_dataset/0007079_flo.jpg} &
\includegraphics[width=.11\linewidth]{figures/supp_dataset/0000496.jpg} &
\includegraphics[width=.11\linewidth]{figures/supp_dataset/0000496_d.jpg} &
\includegraphics[width=.11\linewidth]{figures/supp_dataset/0000496_flo.jpg} \\ 
\includegraphics[width=.11\linewidth]{figures/supp_dataset/0006582.jpg} &
\includegraphics[width=.11\linewidth]{figures/supp_dataset/0006582_d.jpg} &
\includegraphics[width=.11\linewidth]{figures/supp_dataset/0006582_flo.jpg} &
\includegraphics[width=.11\linewidth]{figures/supp_dataset/0002188.jpg} &
\includegraphics[width=.11\linewidth]{figures/supp_dataset/0002188_d.jpg} &
\includegraphics[width=.11\linewidth]{figures/supp_dataset/0002188_flo.jpg} &
\includegraphics[width=.11\linewidth]{figures/supp_dataset/0004772.jpg} &
\includegraphics[width=.11\linewidth]{figures/supp_dataset/0004772_d.jpg} &
\includegraphics[width=.11\linewidth]{figures/supp_dataset/0004772_flo.jpg} \\ 
\includegraphics[width=.11\linewidth]{figures/supp_dataset/0002387.jpg} &
\includegraphics[width=.11\linewidth]{figures/supp_dataset/0002387_d.jpg} &
\includegraphics[width=.11\linewidth]{figures/supp_dataset/0002387_flo.jpg} &
\includegraphics[width=.11\linewidth]{figures/supp_dataset/0001652.jpg} &
\includegraphics[width=.11\linewidth]{figures/supp_dataset/0001652_d.jpg} &
\includegraphics[width=.11\linewidth]{figures/supp_dataset/0001652_flo.jpg} &
\includegraphics[width=.11\linewidth]{figures/supp_dataset/0008075.jpg} &
\includegraphics[width=.11\linewidth]{figures/supp_dataset/0008075_d.jpg} &
\includegraphics[width=.11\linewidth]{figures/supp_dataset/0008075_flo.jpg} \\ 
\includegraphics[width=.11\linewidth]{figures/supp_dataset/0003050.jpg} &
\includegraphics[width=.11\linewidth]{figures/supp_dataset/0003050_d.jpg} &
\includegraphics[width=.11\linewidth]{figures/supp_dataset/0003050_flo.jpg} &
\includegraphics[width=.11\linewidth]{figures/supp_dataset/0004217.jpg} &
\includegraphics[width=.11\linewidth]{figures/supp_dataset/0004217_d.jpg} &
\includegraphics[width=.11\linewidth]{figures/supp_dataset/0004217_flo.jpg} &
\includegraphics[width=.11\linewidth]{figures/supp_dataset/0003902.jpg} &
\includegraphics[width=.11\linewidth]{figures/supp_dataset/0003902_d.jpg} &
\includegraphics[width=.11\linewidth]{figures/supp_dataset/0003902_flo.jpg} \\ 
\includegraphics[width=.11\linewidth]{figures/supp_dataset/0001050.jpg} &
\includegraphics[width=.11\linewidth]{figures/supp_dataset/0001050_d.jpg} &
\includegraphics[width=.11\linewidth]{figures/supp_dataset/0001050_flo.jpg} &
\includegraphics[width=.11\linewidth]{figures/supp_dataset/0003067.jpg} &
\includegraphics[width=.11\linewidth]{figures/supp_dataset/0003067_d.jpg} &
\includegraphics[width=.11\linewidth]{figures/supp_dataset/0003067_flo.jpg} &
\includegraphics[width=.11\linewidth]{figures/supp_dataset/0001105.jpg} &
\includegraphics[width=.11\linewidth]{figures/supp_dataset/0001105_d.jpg} &
\includegraphics[width=.11\linewidth]{figures/supp_dataset/0001105_flo.jpg} \\ 
\includegraphics[width=.11\linewidth]{figures/supp_dataset/0004615.jpg} &
\includegraphics[width=.11\linewidth]{figures/supp_dataset/0004615_d.jpg} &
\includegraphics[width=.11\linewidth]{figures/supp_dataset/0004615_flo.jpg} &
\includegraphics[width=.11\linewidth]{figures/supp_dataset/0007190.jpg} &
\includegraphics[width=.11\linewidth]{figures/supp_dataset/0007190_d.jpg} &
\includegraphics[width=.11\linewidth]{figures/supp_dataset/0007190_flo.jpg} &
\includegraphics[width=.11\linewidth]{figures/supp_dataset/0002453.jpg} &
\includegraphics[width=.11\linewidth]{figures/supp_dataset/0002453_d.jpg} &
\includegraphics[width=.11\linewidth]{figures/supp_dataset/0002453_flo.jpg} \\ 
\includegraphics[width=.11\linewidth]{figures/supp_dataset/0002419.jpg} &
\includegraphics[width=.11\linewidth]{figures/supp_dataset/0002419_d.jpg} &
\includegraphics[width=.11\linewidth]{figures/supp_dataset/0002419_flo.jpg} &
\includegraphics[width=.11\linewidth]{figures/supp_dataset/0002209.jpg} &
\includegraphics[width=.11\linewidth]{figures/supp_dataset/0002209_d.jpg} &
\includegraphics[width=.11\linewidth]{figures/supp_dataset/0002209_flo.jpg} &
\includegraphics[width=.11\linewidth]{figures/supp_dataset/0005310.jpg} &
\includegraphics[width=.11\linewidth]{figures/supp_dataset/0005310_d.jpg} &
\includegraphics[width=.11\linewidth]{figures/supp_dataset/0005310_flo.jpg} \\ 
\includegraphics[width=.11\linewidth]{figures/supp_dataset/0001594.jpg} &
\includegraphics[width=.11\linewidth]{figures/supp_dataset/0001594_d.jpg} &
\includegraphics[width=.11\linewidth]{figures/supp_dataset/0001594_flo.jpg} &
\includegraphics[width=.11\linewidth]{figures/supp_dataset/0001608.jpg} &
\includegraphics[width=.11\linewidth]{figures/supp_dataset/0001608_d.jpg} &
\includegraphics[width=.11\linewidth]{figures/supp_dataset/0001608_flo.jpg} &
\includegraphics[width=.11\linewidth]{figures/supp_dataset/0001322.jpg} &
\includegraphics[width=.11\linewidth]{figures/supp_dataset/0001322_d.jpg} &
\includegraphics[width=.11\linewidth]{figures/supp_dataset/0001322_flo.jpg} \\ 
\includegraphics[width=.11\linewidth]{figures/supp_dataset/0000247.jpg} &
\includegraphics[width=.11\linewidth]{figures/supp_dataset/0000247_d.jpg} &
\includegraphics[width=.11\linewidth]{figures/supp_dataset/0000247_flo.jpg} &
\includegraphics[width=.11\linewidth]{figures/supp_dataset/0001865.jpg} &
\includegraphics[width=.11\linewidth]{figures/supp_dataset/0001865_d.jpg} &
\includegraphics[width=.11\linewidth]{figures/supp_dataset/0001865_flo.jpg} &
\includegraphics[width=.11\linewidth]{figures/supp_dataset/0015729.jpg} &
\includegraphics[width=.11\linewidth]{figures/supp_dataset/0015729_d.jpg} &
\includegraphics[width=.11\linewidth]{figures/supp_dataset/0015729_flo.jpg} \\ 
\includegraphics[width=.11\linewidth]{figures/supp_dataset/0002292.jpg} &
\includegraphics[width=.11\linewidth]{figures/supp_dataset/0002292_d.jpg} &
\includegraphics[width=.11\linewidth]{figures/supp_dataset/0002292_flo.jpg} &
\includegraphics[width=.11\linewidth]{figures/supp_dataset/0009651.jpg} &
\includegraphics[width=.11\linewidth]{figures/supp_dataset/0009651_d.jpg} &
\includegraphics[width=.11\linewidth]{figures/supp_dataset/0009651_flo.jpg} &
\includegraphics[width=.11\linewidth]{figures/supp_dataset/0000050.jpg} &
\includegraphics[width=.11\linewidth]{figures/supp_dataset/0000050_d.jpg} &
\includegraphics[width=.11\linewidth]{figures/supp_dataset/0000050_flo.jpg} \\ 
\includegraphics[width=.11\linewidth]{figures/supp_dataset/0000448.jpg} &
\includegraphics[width=.11\linewidth]{figures/supp_dataset/0000448_d.jpg} &
\includegraphics[width=.11\linewidth]{figures/supp_dataset/0000448_flo.jpg} &
\includegraphics[width=.11\linewidth]{figures/supp_dataset/0004029.jpg} &
\includegraphics[width=.11\linewidth]{figures/supp_dataset/0004029_d.jpg} &
\includegraphics[width=.11\linewidth]{figures/supp_dataset/0004029_flo.jpg} &
\includegraphics[width=.11\linewidth]{figures/supp_dataset/0021889.jpg} &
\includegraphics[width=.11\linewidth]{figures/supp_dataset/0021889_d.jpg} &
\includegraphics[width=.11\linewidth]{figures/supp_dataset/0021889_flo.jpg} \\ 
\includegraphics[width=.11\linewidth]{figures/supp_dataset/0001509.jpg} &
\includegraphics[width=.11\linewidth]{figures/supp_dataset/0001509_d.jpg} &
\includegraphics[width=.11\linewidth]{figures/supp_dataset/0001509_flo.jpg} &
\includegraphics[width=.11\linewidth]{figures/supp_dataset/0001100.jpg} &
\includegraphics[width=.11\linewidth]{figures/supp_dataset/0001100_d.jpg} &
\includegraphics[width=.11\linewidth]{figures/supp_dataset/0001100_flo.jpg} &
\includegraphics[width=.11\linewidth]{figures/supp_dataset/0003369.jpg} &
\includegraphics[width=.11\linewidth]{figures/supp_dataset/0003369_d.jpg} &
\includegraphics[width=.11\linewidth]{figures/supp_dataset/0003369_flo.jpg} \\ 
\includegraphics[width=.11\linewidth]{figures/supp_dataset/0003475.jpg} &
\includegraphics[width=.11\linewidth]{figures/supp_dataset/0003475_d.jpg} &
\includegraphics[width=.11\linewidth]{figures/supp_dataset/0003475_flo.jpg} &
\includegraphics[width=.11\linewidth]{figures/supp_dataset/0000702.jpg} &
\includegraphics[width=.11\linewidth]{figures/supp_dataset/0000702_d.jpg} &
\includegraphics[width=.11\linewidth]{figures/supp_dataset/0000702_flo.jpg} &
\includegraphics[width=.11\linewidth]{figures/supp_dataset/0001052.jpg} &
\includegraphics[width=.11\linewidth]{figures/supp_dataset/0001052_d.jpg} &
\includegraphics[width=.11\linewidth]{figures/supp_dataset/0001052_flo.jpg} \\ 
\includegraphics[width=.11\linewidth]{figures/supp_dataset/0001702.jpg} &
\includegraphics[width=.11\linewidth]{figures/supp_dataset/0001702_d.jpg} &
\includegraphics[width=.11\linewidth]{figures/supp_dataset/0001702_flo.jpg} &
\includegraphics[width=.11\linewidth]{figures/supp_dataset/0001627.jpg} &
\includegraphics[width=.11\linewidth]{figures/supp_dataset/0001627_d.jpg} &
\includegraphics[width=.11\linewidth]{figures/supp_dataset/0001627_flo.jpg} &
\includegraphics[width=.11\linewidth]{figures/supp_dataset/0005126.jpg} &
\includegraphics[width=.11\linewidth]{figures/supp_dataset/0005126_d.jpg} &
\includegraphics[width=.11\linewidth]{figures/supp_dataset/0005126_flo.jpg} \\ 
\includegraphics[width=.11\linewidth]{figures/supp_dataset/0001591.jpg} &
\includegraphics[width=.11\linewidth]{figures/supp_dataset/0001591_d.jpg} &
\includegraphics[width=.11\linewidth]{figures/supp_dataset/0001591_flo.jpg} &
\includegraphics[width=.11\linewidth]{figures/supp_dataset/0001181.jpg} &
\includegraphics[width=.11\linewidth]{figures/supp_dataset/0001181_d.jpg} &
\includegraphics[width=.11\linewidth]{figures/supp_dataset/0001181_flo.jpg} &
\includegraphics[width=.11\linewidth]{figures/supp_dataset/0001812.jpg} &
\includegraphics[width=.11\linewidth]{figures/supp_dataset/0001812_d.jpg} &
\includegraphics[width=.11\linewidth]{figures/supp_dataset/0001812_flo.jpg} \\ 
\includegraphics[width=.11\linewidth]{figures/supp_dataset/0001070.jpg} &
\includegraphics[width=.11\linewidth]{figures/supp_dataset/0001070_d.jpg} &
\includegraphics[width=.11\linewidth]{figures/supp_dataset/0001070_flo.jpg} &
\includegraphics[width=.11\linewidth]{figures/supp_dataset/0003109.jpg} &
\includegraphics[width=.11\linewidth]{figures/supp_dataset/0003109_d.jpg} &
\includegraphics[width=.11\linewidth]{figures/supp_dataset/0003109_flo.jpg} &
\includegraphics[width=.11\linewidth]{figures/supp_dataset/0044410.jpg} &
\includegraphics[width=.11\linewidth]{figures/supp_dataset/0044410_d.jpg} &
\includegraphics[width=.11\linewidth]{figures/supp_dataset/0044410_flo.jpg} \\
Image & Disparity & Flow & Image & Disparity & Flow & Image & Disparity & Flow \\
\end{tabular}
\caption{ Selection of frames from \dataset{} showing diverse scenes, objects and motions.}
\label{fig:supp_dataset}
\end{figure*}

\begin{figure*}[t]
    \centering
    \begin{tabular}{*{6}{c@{\hspace{2px}}}}
    Input & Input flow  & MegaDepth~\cite{li2018megadepth} & Ours ($I_t$) & Ours ($I_t$+$I_{t+1}$+ flow) & GT\\
 	 \includegraphics[width=.16\linewidth]{figures/supp_yt3d_rt/00344_input.jpg}&
  	 \includegraphics[width=.16\linewidth]{figures/supp_yt3d_rt/00344_flo.jpg}&
  	 \includegraphics[width=.16\linewidth]{figures/supp_yt3d_rt/00344_megadepth.jpg}&
  	 \includegraphics[width=.16\linewidth]{figures/supp_yt3d_rt/00344_1frame.jpg}&
  	 \includegraphics[width=.16\linewidth]{figures/supp_yt3d_rt/00344_2frame_consist.jpg}&
  	 \includegraphics[width=.16\linewidth]{figures/supp_yt3d_rt/00344_gt.jpg}\\
 	 \includegraphics[width=.16\linewidth]{figures/supp_yt3d_rt/00246_input.jpg}&
  	 \includegraphics[width=.16\linewidth]{figures/supp_yt3d_rt/00246_flo.jpg}&
  	 \includegraphics[width=.16\linewidth]{figures/supp_yt3d_rt/00246_megadepth.jpg}&
  	 \includegraphics[width=.16\linewidth]{figures/supp_yt3d_rt/00246_1frame.jpg}&
  	 \includegraphics[width=.16\linewidth]{figures/supp_yt3d_rt/00246_2frame_consist.jpg}&
  	 \includegraphics[width=.16\linewidth]{figures/supp_yt3d_rt/00246_gt.jpg}\\
 	 \includegraphics[width=.16\linewidth]{figures/supp_yt3d_rt/00269_input.jpg}&
  	 \includegraphics[width=.16\linewidth]{figures/supp_yt3d_rt/00269_flo.jpg}&
  	 \includegraphics[width=.16\linewidth]{figures/supp_yt3d_rt/00269_megadepth.jpg}&
  	 \includegraphics[width=.16\linewidth]{figures/supp_yt3d_rt/00269_1frame.jpg}&
  	 \includegraphics[width=.16\linewidth]{figures/supp_yt3d_rt/00269_2frame_consist.jpg}&
  	 \includegraphics[width=.16\linewidth]{figures/supp_yt3d_rt/00269_gt.jpg}\\
  	 \includegraphics[width=.16\linewidth]{figures/supp_yt3d_rt/00093_input.jpg}&
  	 \includegraphics[width=.16\linewidth]{figures/supp_yt3d_rt/00093_flo.jpg}&
  	 \includegraphics[width=.16\linewidth]{figures/supp_yt3d_rt/00093_megadepth.jpg}&
  	 \includegraphics[width=.16\linewidth]{figures/supp_yt3d_rt/00093_1frame.jpg}&
  	 \includegraphics[width=.16\linewidth]{figures/supp_yt3d_rt/00093_2frame_consist.jpg}&
  	 \includegraphics[width=.16\linewidth]{figures/supp_yt3d_rt/00093_gt.jpg}\\
  	 \includegraphics[width=.16\linewidth]{figures/supp_yt3d_rt/00081_input.jpg}&
  	 \includegraphics[width=.16\linewidth]{figures/supp_yt3d_rt/00081_flo.jpg}&
  	 \includegraphics[width=.16\linewidth]{figures/supp_yt3d_rt/00081_megadepth.jpg}&
  	 \includegraphics[width=.16\linewidth]{figures/supp_yt3d_rt/00081_1frame.jpg}&
  	 \includegraphics[width=.16\linewidth]{figures/supp_yt3d_rt/00081_2frame_consist.jpg}&
  	 \includegraphics[width=.16\linewidth]{figures/supp_yt3d_rt/00081_gt.jpg}\\
  	 \includegraphics[width=.16\linewidth]{figures/supp_yt3d_rt/00095_input.jpg}&
  	 \includegraphics[width=.16\linewidth]{figures/supp_yt3d_rt/00095_flo.jpg}&
  	 \includegraphics[width=.16\linewidth]{figures/supp_yt3d_rt/00095_megadepth.jpg}&
  	 \includegraphics[width=.16\linewidth]{figures/supp_yt3d_rt/00095_1frame.jpg}&
  	 \includegraphics[width=.16\linewidth]{figures/supp_yt3d_rt/00095_2frame_consist.jpg}&
  	 \includegraphics[width=.16\linewidth]{figures/supp_yt3d_rt/00095_gt.jpg}\\
  	 \includegraphics[width=.16\linewidth]{figures/supp_yt3d_rt/00306_input.jpg}&
  	 \includegraphics[width=.16\linewidth]{figures/supp_yt3d_rt/00306_flo.jpg}&
  	 \includegraphics[width=.16\linewidth]{figures/supp_yt3d_rt/00306_megadepth.jpg}&
  	 \includegraphics[width=.16\linewidth]{figures/supp_yt3d_rt/00306_1frame.jpg}&
  	 \includegraphics[width=.16\linewidth]{figures/supp_yt3d_rt/00306_2frame_consist.jpg}&
  	 \includegraphics[width=.16\linewidth]{figures/supp_yt3d_rt/00306_gt.jpg}\\
  	 \includegraphics[width=.16\linewidth]{figures/supp_yt3d_rt/00545_input.jpg}&
  	 \includegraphics[width=.16\linewidth]{figures/supp_yt3d_rt/00545_flo.jpg}&
  	 \includegraphics[width=.16\linewidth]{figures/supp_yt3d_rt/00545_megadepth.jpg}&
  	 \includegraphics[width=.16\linewidth]{figures/supp_yt3d_rt/00545_1frame.jpg}&
  	 \includegraphics[width=.16\linewidth]{figures/supp_yt3d_rt/00545_2frame_consist.jpg}&
  	 \includegraphics[width=.16\linewidth]{figures/supp_yt3d_rt/00545_gt.jpg}\\
  	 \includegraphics[width=.16\linewidth]{figures/supp_yt3d_rt/00213_input.jpg}&
  	 \includegraphics[width=.16\linewidth]{figures/supp_yt3d_rt/00213_flo.jpg}&
  	 \includegraphics[width=.16\linewidth]{figures/supp_yt3d_rt/00213_megadepth.jpg}&
  	 \includegraphics[width=.16\linewidth]{figures/supp_yt3d_rt/00213_1frame.jpg}&
  	 \includegraphics[width=.16\linewidth]{figures/supp_yt3d_rt/00213_2frame_consist.jpg}&
  	 \includegraphics[width=.16\linewidth]{figures/supp_yt3d_rt/00213_gt.jpg}\\
  	 \includegraphics[width=.16\linewidth]{figures/supp_yt3d_rt/00482_input.jpg}&
  	 \includegraphics[width=.16\linewidth]{figures/supp_yt3d_rt/00482_flo.jpg}&
  	 \includegraphics[width=.16\linewidth]{figures/supp_yt3d_rt/00482_megadepth.jpg}&
  	 \includegraphics[width=.16\linewidth]{figures/supp_yt3d_rt/00482_1frame.jpg}&
  	 \includegraphics[width=.16\linewidth]{figures/supp_yt3d_rt/00482_2frame_consist.jpg}&
  	 \includegraphics[width=.16\linewidth]{figures/supp_yt3d_rt/00482_gt.jpg}\\
   	 \includegraphics[width=.16\linewidth]{figures/supp_yt3d_rt/00429_input.jpg}&
  	 \includegraphics[width=.16\linewidth]{figures/supp_yt3d_rt/00429_flo.jpg}&
  	 \includegraphics[width=.16\linewidth]{figures/supp_yt3d_rt/00429_megadepth.jpg}&
  	 \includegraphics[width=.16\linewidth]{figures/supp_yt3d_rt/00429_1frame.jpg}&
  	 \includegraphics[width=.16\linewidth]{figures/supp_yt3d_rt/00429_2frame_consist.jpg}&
  	 \includegraphics[width=.16\linewidth]{figures/supp_yt3d_rt/00429_gt.jpg}\\
   	 \includegraphics[width=.16\linewidth]{figures/supp_yt3d_rt/00380_input.jpg}&
  	 \includegraphics[width=.16\linewidth]{figures/supp_yt3d_rt/00380_flo.jpg}&
  	 \includegraphics[width=.16\linewidth]{figures/supp_yt3d_rt/00380_megadepth.jpg}&
  	 \includegraphics[width=.16\linewidth]{figures/supp_yt3d_rt/00380_1frame.jpg}&
  	 \includegraphics[width=.16\linewidth]{figures/supp_yt3d_rt/00380_2frame_consist.jpg}&
  	 \includegraphics[width=.16\linewidth]{figures/supp_yt3d_rt/00380_gt.jpg}\\
  	 \includegraphics[width=.16\linewidth]{figures/supp_yt3d_rt/00377_input.jpg}&
  	 \includegraphics[width=.16\linewidth]{figures/supp_yt3d_rt/00377_flo.jpg}&
  	 \includegraphics[width=.16\linewidth]{figures/supp_yt3d_rt/00377_megadepth.jpg}&
  	 \includegraphics[width=.16\linewidth]{figures/supp_yt3d_rt/00377_1frame.jpg}&
  	 \includegraphics[width=.16\linewidth]{figures/supp_yt3d_rt/00377_2frame_consist.jpg}&
  	 \includegraphics[width=.16\linewidth]{figures/supp_yt3d_rt/00377_gt.jpg}\\
   	 \includegraphics[width=.16\linewidth]{figures/supp_yt3d_rt/00334_input.jpg}&
  	 \includegraphics[width=.16\linewidth]{figures/supp_yt3d_rt/00334_flo.jpg}&
  	 \includegraphics[width=.16\linewidth]{figures/supp_yt3d_rt/00334_megadepth.jpg}&
  	 \includegraphics[width=.16\linewidth]{figures/supp_yt3d_rt/00334_1frame.jpg}&
  	 \includegraphics[width=.16\linewidth]{figures/supp_yt3d_rt/00334_2frame_consist.jpg}&
  	 \includegraphics[width=.16\linewidth]{figures/supp_yt3d_rt/00334_gt.jpg}\\
    Input & Input flow  & MegaDepth & Ours ($I_t$) & Ours ($I_t$+$I_{t+1}$+ flow) & GT\\
    \end{tabular}
    \caption{Qualitative results of comparing MegaDepth vs baseline monocular depth prediction vs Ours.}
    \label{fig:supp_yt3d_comp}
\end{figure*}

\begin{figure*}[ht]
    \centering
    \begin{tabular}{*{6}{c@{\hspace{2px}}}}
    \includegraphics[width=.16\linewidth]{figures/supp_kitti_rt/00558_input.jpg} &
    \includegraphics[width=.16\linewidth]{figures/supp_kitti_rt/00558_flo.jpg} &
    \includegraphics[width=.16\linewidth]{figures/supp_kitti_rt/00558_megadepth.jpg} &
    \includegraphics[width=.16\linewidth]{figures/supp_kitti_rt/00558_1frame.jpg} &
    \includegraphics[width=.16\linewidth]{figures/supp_kitti_rt/00558_2frame_consist.jpg} &
    \includegraphics[width=.16\linewidth]{figures/supp_kitti_rt/00558.jpg}\\
    \includegraphics[width=.16\linewidth]{figures/supp_kitti_rt/00564_input.jpg} &
    \includegraphics[width=.16\linewidth]{figures/supp_kitti_rt/00564_flo.jpg} &
    \includegraphics[width=.16\linewidth]{figures/supp_kitti_rt/00564_megadepth.jpg} &
    \includegraphics[width=.16\linewidth]{figures/supp_kitti_rt/00564_1frame.jpg} &
    \includegraphics[width=.16\linewidth]{figures/supp_kitti_rt/00564_2frame_consist.jpg} &
    \includegraphics[width=.16\linewidth]{figures/supp_kitti_rt/00564.jpg}\\
    \includegraphics[width=.16\linewidth]{figures/supp_kitti_rt/00540_input.jpg} &
    \includegraphics[width=.16\linewidth]{figures/supp_kitti_rt/00540_flo.jpg} &
    \includegraphics[width=.16\linewidth]{figures/supp_kitti_rt/00540_megadepth.jpg} &
    \includegraphics[width=.16\linewidth]{figures/supp_kitti_rt/00540_1frame.jpg} &
    \includegraphics[width=.16\linewidth]{figures/supp_kitti_rt/00540_2frame_consist.jpg} &
    \includegraphics[width=.16\linewidth]{figures/supp_kitti_rt/00540.jpg}\\
    \includegraphics[width=.16\linewidth]{figures/supp_kitti_rt/00563_input.jpg} &
    \includegraphics[width=.16\linewidth]{figures/supp_kitti_rt/00563_flo.jpg} &
    \includegraphics[width=.16\linewidth]{figures/supp_kitti_rt/00563_megadepth.jpg} &
    \includegraphics[width=.16\linewidth]{figures/supp_kitti_rt/00563_1frame.jpg} &
    \includegraphics[width=.16\linewidth]{figures/supp_kitti_rt/00563_2frame_consist.jpg}&
    \includegraphics[width=.16\linewidth]{figures/supp_kitti_rt/00563.jpg}\\
    Input & Input flow & MegaDepth & Ours ($I_t$) & Ours ($I_t$+$I_{t+1}$+ flow) & GT\\
    \end{tabular}
    \caption{Qualitative results on KITTI}
    \label{fig:supp_kitti_comp}
\end{figure*}

\include{file}

%% file: main.bbl
\begin{thebibliography}{10}\itemsep=-1pt

\bibitem{pyscendetect}
Pyscendetect.
\newblock \url{https://pyscenedetect.readthedocs.io}, 2008.
\newblock [Online; accessed 1-July-2018].

\bibitem{akhter2009nonrigid}
I.~Akhter, Y.~Sheikh, S.~Khan, and T.~Kanade.
\newblock Nonrigid structure from motion in trajectory space.
\newblock In {\em Advances in neural information processing systems}, pages
  41--48, 2009.

\bibitem{bregler2000recovering}
C.~Bregler, A.~Hertzmann, and H.~Biermann.
\newblock Recovering non-rigid 3d shape from image streams.
\newblock In {\em Computer Vision and Pattern Recognition, 2000. Proceedings.
  IEEE Conference on}, volume~2, pages 690--696. IEEE, 2000.

\bibitem{diw}
W.~Chen, Z.~Fu, D.~Yang, and J.~Deng.
\newblock Single-image depth perception in the wild.
\newblock In {\em Advances in Neural Information Processing Systems}, pages
  730--738, 2016.

\bibitem{dai2014simple}
Y.~Dai, H.~Li, and M.~He.
\newblock A simple prior-free method for non-rigid structure-from-motion
  factorization.
\newblock {\em International Journal of Computer Vision}, 107(2):101--122,
  2014.

\bibitem{eigen2014depth}
D.~Eigen, C.~Puhrsch, and R.~Fergus.
\newblock Depth map prediction from a single image using a multi-scale deep
  network.
\newblock In {\em Advances in neural information processing systems}, pages
  2366--2374, 2014.

\bibitem{garg2016unsupervised}
R.~Garg, V.~K. BG, G.~Carneiro, and I.~Reid.
\newblock Unsupervised cnn for single view depth estimation: Geometry to the
  rescue.
\newblock In {\em European Conference on Computer Vision}, pages 740--756.
  Springer, 2016.

\bibitem{geiger2012we}
A.~Geiger, P.~Lenz, and R.~Urtasun.
\newblock Are we ready for autonomous driving? the kitti vision benchmark
  suite.
\newblock In {\em Computer Vision and Pattern Recognition (CVPR), 2012 IEEE
  Conference on}, pages 3354--3361. IEEE, 2012.

\bibitem{godard2017unsupervised}
C.~Godard, O.~Mac~Aodha, and G.~J. Brostow.
\newblock Unsupervised monocular depth estimation with left-right consistency.
\newblock In {\em CVPR}, volume~2, page~7, 2017.

\bibitem{Guo_2018_ECCV}
X.~Guo, H.~Li, S.~Yi, J.~Ren, and X.~Wang.
\newblock Learning monocular depth by distilling cross-domain stereo networks.
\newblock In {\em The European Conference on Computer Vision (ECCV)}, September
  2018.

\bibitem{He2017MaskR}
K.~He, G.~Gkioxari, P.~Doll{\'a}r, and R.~B. Girshick.
\newblock Mask r-cnn.
\newblock {\em 2017 IEEE International Conference on Computer Vision (ICCV)},
  pages 2980--2988, 2017.

\bibitem{huang2018deepmvs}
P.-H. Huang, K.~Matzen, J.~Kopf, N.~Ahuja, and J.-B. Huang.
\newblock Deepmvs: Learning multi-view stereopsis.
\newblock In {\em Proceedings of the IEEE Conference on Computer Vision and
  Pattern Recognition}, pages 2821--2830, 2018.

\bibitem{ilg2017flownet}
E.~Ilg, N.~Mayer, T.~Saikia, M.~Keuper, A.~Dosovitskiy, and T.~Brox.
\newblock Flownet 2.0: Evolution of optical flow estimation with deep networks.
\newblock In {\em IEEE conference on computer vision and pattern recognition
  (CVPR)}, volume~2, page~6, 2017.

\bibitem{kong2016prior}
C.~Kong and S.~Lucey.
\newblock Prior-less compressible structure from motion.
\newblock In {\em Proceedings of the IEEE Conference on Computer Vision and
  Pattern Recognition}, pages 4123--4131, 2016.

\bibitem{openimages}
I.~Krasin, T.~Duerig, N.~Alldrin, V.~Ferrari, S.~Abu-El-Haija, A.~Kuznetsova,
  H.~Rom, J.~Uijlings, S.~Popov, A.~Veit, S.~Belongie, V.~Gomes, A.~Gupta,
  C.~Sun, G.~Chechik, D.~Cai, Z.~Feng, D.~Narayanan, and K.~Murphy.
\newblock Openimages: A public dataset for large-scale multi-label and
  multi-class image classification.
\newblock {\em Dataset available from https://github.com/openimages}, 2017.

\bibitem{kumar2018scalable}
S.~Kumar, A.~Cherian, Y.~Dai, and H.~Li.
\newblock Scalable dense non-rigid structure-from-motion: A grassmannian
  perspective.
\newblock {\em CVPR}, 2018.

\bibitem{kumar2017monocular}
S.~Kumar, Y.~Dai, and H.~Li.
\newblock Monocular dense 3d reconstruction of a complex dynamic scene from two
  perspective frames.
\newblock In {\em IEEE International Conference on Computer Vision}, pages
  4649--4657, 2017.

\bibitem{kuznietsov2017semi}
Y.~Kuznietsov, J.~St{\"u}ckler, and B.~Leibe.
\newblock Semi-supervised deep learning for monocular depth map prediction.
\newblock In {\em Proc. of the IEEE Conference on Computer Vision and Pattern
  Recognition}, pages 6647--6655, 2017.

\bibitem{laina2016deeper}
I.~Laina, C.~Rupprecht, V.~Belagiannis, F.~Tombari, and N.~Navab.
\newblock Deeper depth prediction with fully convolutional residual networks.
\newblock In {\em 3D Vision (3DV), 2016 Fourth International Conference on},
  pages 239--248. IEEE, 2016.

\bibitem{lee2013procrustean}
M.~Lee, J.~Cho, C.-H. Choi, and S.~Oh.
\newblock Procrustean normal distribution for non-rigid structure from motion.
\newblock In {\em Proceedings of the IEEE Conference on Computer Vision and
  Pattern Recognition}, pages 1280--1287, 2013.

\bibitem{li2018megadepth}
Z.~Li and N.~Snavely.
\newblock Megadepth: Learning single-view depth prediction from internet
  photos.
\newblock In {\em Proceedings of the IEEE Conference on Computer Vision and
  Pattern Recognition}, pages 2041--2050, 2018.

\bibitem{deeplens2018}
W.~Lijun, S.~Xiaohui, Z.~Jianming, W.~Oliver, H.~Chih-Yao, K.~Sarah, and
  L.~Huchuan.
\newblock Deeplens: Shallow depth of field from a single image.
\newblock {\em ACM Trans. Graph. (Proc. SIGGRAPH Asia)}, 37(6):6:1--6:11, 2018.

\bibitem{MahjourianWA17}
R.~Mahjourian, M.~Wicke, and A.~Angelova.
\newblock Geometry-based next frame prediction from monocular video.
\newblock In {\em {IEEE} Intelligent Vehicles Symposium, {IV} 2017, Los
  Angeles, CA, USA, June 11-14, 2017}, pages 1700--1707, 2017.

\bibitem{mahjourian2018unsupervised}
R.~Mahjourian, M.~Wicke, and A.~Angelova.
\newblock Unsupervised learning of depth and ego-motion from monocular video
  using 3d geometric constraints.
\newblock In {\em Proceedings of the IEEE Conference on Computer Vision and
  Pattern Recognition}, pages 5667--5675, 2018.

\bibitem{mur2015orb}
R.~Mur-Artal, J.~M.~M. Montiel, and J.~D. Tardos.
\newblock Orb-slam: a versatile and accurate monocular slam system.
\newblock {\em IEEE Transactions on Robotics}, 31(5):1147--1163, 2015.

\bibitem{nyu_v2}
P.~K. Nathan~Silberman, Derek~Hoiem and R.~Fergus.
\newblock Indoor segmentation and support inference from rgbd images.
\newblock In {\em ECCV}, 2012.

\bibitem{dmde}
R.~Ranftl, V.~Vineet, Q.~Chen, and V.~Koltun.
\newblock Dense monocular depth estimation in complex dynamic scenes.
\newblock In {\em Proceedings of the IEEE Conference on Computer Vision and
  Pattern Recognition}, pages 4058--4066, 2016.

\bibitem{saxena2009make3d}
A.~Saxena, M.~Sun, and A.~Y. Ng.
\newblock Make3d: Learning 3d scene structure from a single still image.
\newblock {\em IEEE transactions on pattern analysis and machine intelligence},
  31(5):824--840, 2009.

\bibitem{schoenberger2016sfm}
J.~L. Sch\"{o}nberger and J.-M. Frahm.
\newblock Structure-from-motion revisited.
\newblock In {\em Conference on Computer Vision and Pattern Recognition
  (CVPR)}, 2016.

\bibitem{schonberger2016structure}
J.~L. Schonberger and J.-M. Frahm.
\newblock Structure-from-motion revisited.
\newblock In {\em Proceedings of the IEEE Conference on Computer Vision and
  Pattern Recognition}, pages 4104--4113, 2016.

\bibitem{schoenberger2016mvs}
J.~L. Sch\"{o}nberger, E.~Zheng, M.~Pollefeys, and J.-M. Frahm.
\newblock Pixelwise view selection for unstructured multi-view stereo.
\newblock In {\em European Conference on Computer Vision (ECCV)}, 2016.

\bibitem{ummenhofer2017demon}
B.~Ummenhofer, H.~Zhou, J.~Uhrig, N.~Mayer, E.~Ilg, A.~Dosovitskiy, and
  T.~Brox.
\newblock Demon: Depth and motion network for learning monocular stereo.
\newblock In {\em IEEE Conference on computer vision and pattern recognition
  (CVPR)}, volume~5, page~6, 2017.

\bibitem{Wang_2018_CVPR}
C.~Wang, J.~Miguel~Buenaposada, R.~Zhu, and S.~Lucey.
\newblock Learning depth from monocular videos using direct methods.
\newblock In {\em The IEEE Conference on Computer Vision and Pattern
  Recognition (CVPR)}, June 2018.

\bibitem{xian2018monocular}
K.~Xian, C.~Shen, Z.~Cao, H.~Lu, Y.~Xiao, R.~Li, and Z.~Luo.
\newblock Monocular relative depth perception with web stereo data supervision.
\newblock In {\em Proceedings of the IEEE Conference on Computer Vision and
  Pattern Recognition}, pages 311--320, 2018.

\bibitem{xie2016deep3d}
J.~Xie, R.~Girshick, and A.~Farhadi.
\newblock Deep3d: Fully automatic 2d-to-3d video conversion with deep
  convolutional neural networks.
\newblock In {\em European Conference on Computer Vision}, pages 842--857.
  Springer, 2016.

\bibitem{yang2018lego}
Z.~Yang, P.~Wang, Y.~Wang, W.~Xu, and R.~Nevatia.
\newblock Lego: Learning edge with geometry all at once by watching videos.
\newblock In {\em Proceedings of the IEEE Conference on Computer Vision and
  Pattern Recognition}, pages 225--234, 2018.

\bibitem{zhan2018unsupervised}
H.~Zhan, R.~Garg, C.~S. Weerasekera, K.~Li, H.~Agarwal, and I.~Reid.
\newblock Unsupervised learning of monocular depth estimation and visual
  odometry with deep feature reconstruction.
\newblock In {\em Proceedings of the IEEE Conference on Computer Vision and
  Pattern Recognition}, pages 340--349, 2018.

\bibitem{zhou2018deeptam}
H.~Zhou, B.~Ummenhofer, and T.~Brox.
\newblock Deeptam: Deep tracking and mapping.
\newblock In {\em European Conference on Computer Vision (ECCV)}, 2018.

\bibitem{zhou2017unsupervised}
T.~Zhou, M.~Brown, N.~Snavely, and D.~G. Lowe.
\newblock Unsupervised learning of depth and ego-motion from video.
\newblock In {\em CVPR}, volume~2, page~7, 2017.

\bibitem{zhou2018stereo}
T.~Zhou, R.~Tucker, J.~Flynn, G.~Fyffe, and N.~Snavely.
\newblock Stereo magnification: Learning view synthesis using multiplane
  images.
\newblock {\em arXiv preprint arXiv:1805.09817}, 2018.

\bibitem{zhu2014complex}
Y.~Zhu, D.~Huang, F.~De~La~Torre, and S.~Lucey.
\newblock Complex non-rigid motion 3d reconstruction by union of subspaces.
\newblock In {\em Proceedings of the IEEE Conference on Computer Vision and
  Pattern Recognition}, pages 1542--1549, 2014.

\bibitem{zou2018dfnet}
Y.~Zou, Z.~Luo, and J.-B. Huang.
\newblock Df-net: Unsupervised joint learning of depth and flow using
  cross-task consistency.
\newblock In {\em European Conference on Computer Vision}, 2018.

\end{thebibliography}
